\documentclass{article} % For LaTeX2e
\usepackage{iclr2024_conference,times}

\usepackage{amsmath,amsfonts,bm}

\def\eqref#1{equation~\ref{#1}}
\def\1{\bm{1}}

\DeclareMathAlphabet{\mathsfit}{\encodingdefault}{\sfdefault}{m}{sl}
\SetMathAlphabet{\mathsfit}{bold}{\encodingdefault}{\sfdefault}{bx}{n}

\usepackage{hyperref}
\usepackage{url}

\usepackage{graphicx}
\usepackage{booktabs}
\usepackage{multirow}
\usepackage{graphicx}
\usepackage{amssymb}
\usepackage{amsmath}
\usepackage{makecell}
\usepackage{xspace}
\usepackage{arydshln}
\usepackage{makecell}
\usepackage{marvosym}
\usepackage{color,colortbl}
\usepackage{tcolorbox}
\tcbuselibrary{skins}
\usepackage{array}
\usepackage{tabularx}
\tcbuselibrary{breakable}

\usepackage{bbding} 
\usepackage{enumitem}
\usepackage{svg}
\usepackage{subcaption}

\definecolor{myblue}{RGB}{71,135,237}

\title{\textit{VidEgoThink}: Assessing Egocentric Video Understanding Capabilities for Embodied AI}

\author{Sijie Cheng\textsuperscript{1,2,6}\footnotemark[2], Kechen Fang\textsuperscript{2,5*}, YangYang Yu\textsuperscript{2,5*}, Sicheng Zhou\textsuperscript{2,3*}, Bohao Li\textsuperscript{4,6} \\
\textbf{Ye Tian\textsuperscript{6}, Tingguang Li\textsuperscript{6}, Lei Han\textsuperscript{6\Letter}, Yang Liu\textsuperscript{1,2\Letter}}\\
\textsuperscript{1}Department of Computer Science and Technology, Tsinghua University \\
\textsuperscript{2}Institute for AI Industry Research (AIR), Tsinghua University \\
\textsuperscript{3}Department of Mechanical and Industrial Engineering, University of Toronto \\
\textsuperscript{4}School of Data Science, The Chinese University of HongKong\\
\textsuperscript{5}Zhili College, Tsinghua University \quad \textsuperscript{6}Tencent Robotics X  \\
\texttt{csj23@mails.tsinghua.edu.cn} \\
}

\iclrfinalcopy % Uncomment for camera-ready version, but NOT for submission.
\begin{document}

\footnotetext[2]{Project Leader \textsuperscript{*}Equal contribution}
\maketitle
\vspace{-6mm}
\begin{abstract}
Recent advancements in Multi-modal Large Language Models (MLLMs) have opened new avenues for applications in Embodied AI.
Building on previous work, EgoThink, we introduce VidEgoThink, a comprehensive benchmark for evaluating egocentric video understanding capabilities. 
To bridge the gap between MLLMs and low-level control in Embodied AI, we design four key interrelated tasks: \textit{video question-answering}, \textit{hierarchy planning}, \textit{visual grounding} and \textit{reward modeling}. 
To minimize manual annotation costs, we develop an automatic data generation pipeline based on the Ego4D dataset, leveraging the prior knowledge and multimodal capabilities of GPT-4o.
Three human annotators then filter the generated data to ensure diversity and quality, resulting in the VidEgoThink benchmark.
We conduct extensive experiments with three types of models: API-based MLLMs, open-source image-based MLLMs, and open-source video-based MLLMs. 
Experimental results indicate that all MLLMs, including GPT-4o, perform poorly across all tasks related to egocentric video understanding.
These findings suggest that foundation models still require significant advancements to be effectively applied to first-person scenarios in Embodied AI.
In conclusion, VidEgoThink reflects a research trend towards employing MLLMs for egocentric vision, akin to human capabilities, enabling active observation and interaction in the complex real-world environments.

% The abstract paragraph should be indented 1/2~inch (3~picas) on both left and
% right-hand margins. Use 10~point type, with a vertical spacing of 11~points.
% The word \textsc{Abstract} must be centered, in small caps, and in point size 12. Two
% line spaces precede the abstract. The abstract must be limited to one
% paragraph.

\end{abstract}

\begin{figure}[h!]
    \centering
    \vspace{-4mm}
    \includegraphics[width=1\linewidth]{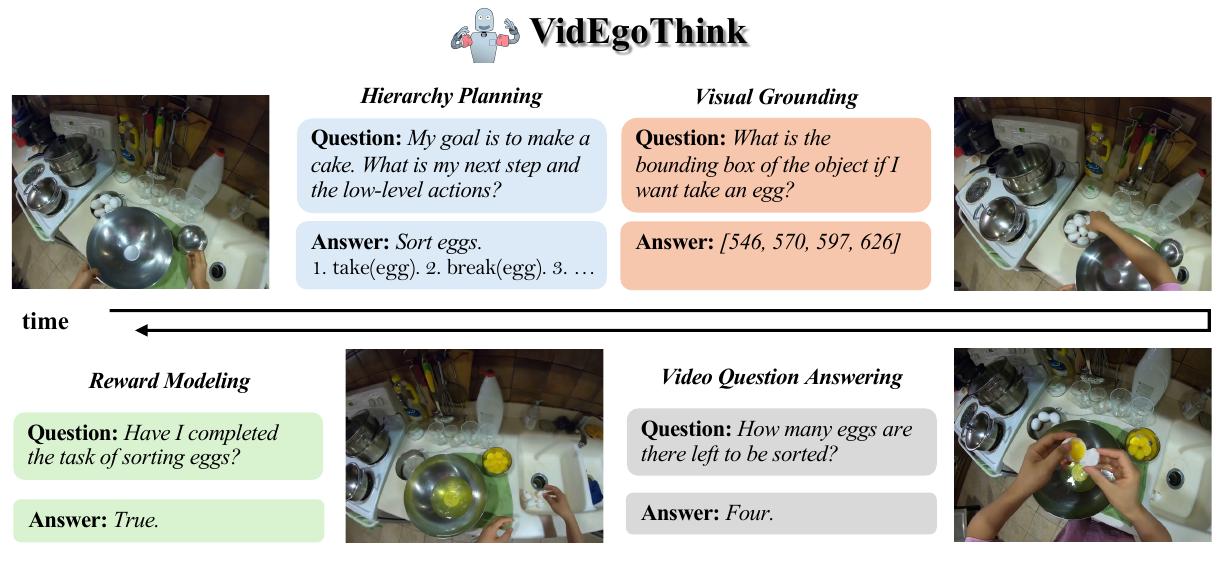}
    \vspace{-7mm}
    \caption{The main tasks of VidEgoThink benchmark to comprehensively assess the egocentric video understanding capabilities in Embodied AI. There are four types of tasks, including \textit{video question answering}, \textit{hierarchy planning}, \textit{visual grounding}, and \textit{reward modeling}. These four tasks are complementary to each other to implement a complete goal for Embodied AI.}
    \label{fig:intro}
\end{figure}

\clearpage

\section{Introduction}

In recent years, Multi-modal Large Language Models (MLLMs;~\citealp{du2022survey, gan2022vision,tang2023video}) have made significant strides in conventional vision-language tasks~\citep{alayrac2022flamingo, driess2023palm, li2023blip}, profoundly impacting the field of Embodied Artificial Intelligence (Embodied AI;~\citealp{ahn2022can, kuo2022f, huang2023voxposer, zitkovich2023rt}). 
Training data~\citep{sharma2018conceptual, schuhmann2022laion, lin2014microsoft, jia2021scaling} for predominate MLLMs are typically collected from object-centric and exocentric perspectives, mirroring the distribution of conventional vision-language benchmarks~\citep{liu2023mmbench, xu2023lvlm, li2023seed, ning2023video}, which focus primarily on object and scene understanding.
However, to be effectively applied in Embodied AI, it is crucial not only to understand the surrounding environment but also to have extensive knowledge about the relationship between ``myself'' and the environment. 
For example, compared to the absolute position in the whole environment (e.g., ``\textit{the microwave is in the kitchen}''), the relative position to my body is more important (e.g., ``\textit{the microwave is one meter to my right}'') for interaction and manipulation.
Therefore, \textbf{egocentric videos}~\citep{grauman2022ego4d, damen2018scaling}, containing observations typical of third-person perspectives and additional interactions with the surrounding environment, can improve predominate MLLMs to be more general and expand their applications to the real world.
% Therefore, \textbf{egocentric videos}~\citep{grauman2022ego4d, damen2018scaling}, containing sufficient information about ``myself'' and the interaction with the surrounding environment in daily-life activities, can improve predominate MLLMs to be more general and expand their applications to the real world.
% Furthermore, it is worth noting that the first-person videos also include observations typical of third-person perspectives and additional interactions with the environment.

Various egocentric benchmarks~\citep{Cheng_2024_CVPR, fan2019egovqa}, as shown in Table~\ref{tab:comparison}, have emerged to evaluate the capabilities of MLLMs from a first-person perspective.
For instance, EgoTaskQA~\citep{jia2022egotaskqa} and EgoPlan~\citep{chen2023egoplan} assess the planning capabilities of MLLMs for long-horizon tasks, while EgoSchema~\citep{mangalam2024egoschema} aims to diagnose the understanding of very long-form video.
However, the absence of a comprehensive video benchmark from the egocentric perspective presents a significant challenge to the development of general foundation models.
Furthermore, current benchmarks, both in task design and textual output forms, focus on traditional video question-answering settings and neglect the potential to support downstream applications in Embodied AI, such as glass devices or autonomous robots.
For example, the natural language output format (e.g., ``\textit{put salmon in microwave}'') cannot be directly processed by robotics to take actions, whereas bounding boxes of grounded objects (e.g., ``\textit{microwave} [\textit{290, 202, 835, 851}]'' or function calls for low-level actions (e.g., ``\texttt{find}(\textit{microwave})'') align more closely with the input requirements of robotic control systems.
Therefore, it is crucial to design suitable task formats that can be effectively applied to downstream applications in Embodied AI.

In this paper, we introduce \textit{VidEgoThink}, as illustrated in Figure~\ref{fig:intro}, a comprehensive egocentric video understanding benchmark aimed at better aligning the capabilities of MLLMs for application in Embodied AI.
Due to the stratospheric demand for training data of end-to-end Vision-Language-Action models~\citep{driess2023palm, padalkar2023open, li2024decisionnce}, systems in Embodied AI are always structured into specialized hierarchical components.
In detail, MLLMs can perform several key functions: 
(1) \textit{video question-answering}, the basic module to comprehend the surrounding environment and human activities, and then generate corresponding responses to specific instructions~\citep{Cheng_2024_CVPR, fan2019egovqa, jia2022egotaskqa};
(2) \textit{hierarchy planning}, the core component to decompose high-level instructions to mid-level sub-goals and low-level actions~\citep{ahn2022can, huang2022inner, huang2022language};
(3) \textit{visual grounding}, the detector module to help Embodied AI system ground complex instruction to the physical world~\citep{gao2023physically, chiang2024mobility, munasinghe2023pg};
(4) \textit{reward modeling}, the auxiliary module to classify task completion and further provide feedback according to the observations~\citep{kwon2023reward, di2023towards, yu2023language}.
Rather than solely considering traditional question-answering or planning tasks like previous egocentric benchmarks, we specifically design these four tasks to comprehensively evaluate the capabilities for different functions of MLLMs in Embodied AI.

\begin{table*}[h!]
    \centering
    \caption{Comparison of recent evaluation benchmarks of multimodal large language models and our proposed benchmark VidEgoThink. VQA/HP/VG/RM indicate visual question answering, hierarchy planning, visual grounding, and reward modeling. Existing/Handcraft/Automatic denote the way of collecting data, including existing dataset, manual annotation, and automatic generation. }
    \resizebox{\linewidth}{!}{
    \begin{tabular}{lccccccccrr}
    \toprule
        \multirow{2}{*}{\textbf{Benchmark}} & \multirow{2}{*}{\makecell[c]{\textbf{Comprehensive} \\ \textbf{Capabilities}}} & \multicolumn{2}{c}{\textbf{View}} & \multicolumn{4}{c}{\textbf{Task Type}} & \multirow{2}{*}{\makecell[c]{\textbf{Data} \\ \textbf{Source}}} & \multirow{2}{*}{\makecell[c]{\textbf{Average} \\ \textbf{Length}}} & \multirow{2}{*}{\makecell[c]{\textbf{Total} \\ \textbf{Size}}} \\
        \cmidrule(rl){3-4}
        \cmidrule(rl){5-8}
        & & \textbf{Observe} & \textbf{Interact} & \textbf{VQA} & \textbf{HP} & \textbf{VG} & \textbf{RM} \\
    \midrule
        ActivityNet-QA & \XSolidBrush & \Checkmark & \XSolidBrush & \Checkmark & \XSolidBrush & \XSolidBrush & \XSolidBrush & Handcraft & 180s & 58,000 \\
        SEED-Bench-2 & \Checkmark & \Checkmark & \XSolidBrush & \Checkmark & \XSolidBrush & \XSolidBrush & \XSolidBrush & Handcraft & - & 24,000 \\
        AutoEval-Video & \Checkmark & \Checkmark & \XSolidBrush & \Checkmark & \XSolidBrush & \XSolidBrush & \XSolidBrush & Handcraft & 14.58s & 327 \\
        Video-Bench & \Checkmark & \Checkmark & \XSolidBrush & \Checkmark & \XSolidBrush & \XSolidBrush & \XSolidBrush & Existing & - & 15,000\\
        Perception Test & \XSolidBrush & \Checkmark & \XSolidBrush & \Checkmark & \XSolidBrush & \Checkmark & \XSolidBrush & Handcraft & 23s & 11,600 \\
        OpenEQA & \XSolidBrush & \Checkmark & \XSolidBrush & \Checkmark & \XSolidBrush & \XSolidBrush & \XSolidBrush & Handcraft & - & 1,600\\
        MVBench & \Checkmark & \Checkmark & \Checkmark & \Checkmark & \XSolidBrush & \XSolidBrush & \XSolidBrush & Existing & (5s, 35s) & 4,000\\
        EgoVQA & \XSolidBrush & \Checkmark & \Checkmark & \Checkmark & \XSolidBrush & \XSolidBrush & \XSolidBrush & Handcraft & (20s, 100s) & 520 \\
        EgoThink & \Checkmark & \XSolidBrush & \Checkmark & \Checkmark & \Checkmark & \XSolidBrush & \XSolidBrush & Handcraft & - & 700 \\
        EgoTaskQA & \XSolidBrush & \XSolidBrush & \Checkmark & \Checkmark & \XSolidBrush & \XSolidBrush & \XSolidBrush & Automatic & 25s & 40,000 \\
        EgoPlan-Bench & \XSolidBrush & \XSolidBrush & \Checkmark & \XSolidBrush & \Checkmark & \XSolidBrush & \XSolidBrush & Automatic & - & 3,400 \\
        EgoSchema & \XSolidBrush & \XSolidBrush & \Checkmark & \Checkmark & \XSolidBrush & \XSolidBrush & \XSolidBrush & Automatic & 180s & 5,000 \\
        \midrule
        \textbf{VidEgoThink} (Ours) & \Checkmark & \Checkmark & \Checkmark & \Checkmark & \Checkmark & \Checkmark & \Checkmark & Automatic & 270.74s & 4,993  \\
    \bottomrule
    \end{tabular}}
    \label{tab:comparison}
\end{table*}

Considering the high cost of manually labeling data for four different tasks, we design a series of automatic construction pipelines leveraging existing annotations from the Ego4D dataset~\citep{grauman2022ego4d}.
we use GPT-4o, known for its superior reasoning capabilities, to generate appropriate question-answering pairs by combining our designed prompts with existing human annotations.
For the reward modeling task, we further adopt clipped images from each video to generate feedback for negative instances.
To ensure diversity and quality, three annotators are asked to filter the automatically generated instances.
For evaluation, we extensively compare 14 MLLMs across three categories: API-based MLLMs, open-source image-based MLLMs, and open-source video-based MLLMs. 
Experimental results indicate that all MLLMs perform poorly across all tasks.
For example, GPT-4o with 32 frames and 8 frames achieve only 31.17\% and 32.83\% accuracy in video question-answering tasks.
Detailed scores reveal that while MLLMs can determine existence across object, action, and scene dimensions, they particularly lack the ability to judge order or sequence.
In other tasks, although GPT-4o's performance is subpar, other open-source MLLMs are almost completely unusable, showing significant performance gaps.
Overall, applying current MLLMs directly to first-person scenarios in Embodied AI remains challenging and requires further effort.
However, MLLMs hold great potential for advancing Egocentric Vision and Embodied AI, offering ample room for exploration and improvement.

\section{Related Work} 

\noindent\textbf{Multi-modal Large Language Models.}
The advancement of large language models (LLMs; \citealp{brown2020language, ouyang2022traininglanguagemodelsfollow, wang2024openchatadvancingopensourcelanguage}) now extend into MLLMs.
Visual modules, such as CLIP~\citep{radford2021learningtransferablevisualmodels} and Q-Former~\citep{dai2024instructblip}, are integrated with pre-trained LLMs using various transition layers, equipping them with visual capabilities. 
From the wide selection of open-source LLMs, numerous image-based MLLMs~\citep{chen2023sharegpt4vimprovinglargemultimodal, liu2024improved, zhang2023llamaadapter, dai2024instructblip, alayrac2022flamingo} have emerged. 
Moreover, the popularization of these image-based MLLMs has driven advancements in video perception.
Video-based models like Video-LLaVA \citep{lin2023video}, Vision-LLaMA \citep{chu2024visionllama}, and PandaGPT\citep{su2023pandagpt} are capable of capturing the temporal information present in video form. 
In this work, we explore egocentric video understanding capabilities of MLLMs.

\noindent\textbf{Video-Langugae Benchmarks.}
Numerous video-language benchmarks assess MLLMs, primarily focusing on instruction-following via visual question-answering tasks~\citep{ning2023video, li2023mvbench, NEURIPS2023_8540fba4}. 
Few benchmarks explore egocentric videos~\citep{mangalam2024egoschema, jia2022egotaskqa}, like EgoTaskQA~\citep{jia2022egotaskqa}, EgoPlan-Bench~\citep{chen2023egoplan}, and EgoGoalStep~\citep{song2023egogoalstep}.
However, they often lack variety in assessed capabilities.
EgoThink~\citep{Cheng_2024_CVPR} covers more comprehensive capabilities but uses static images.
Moreover, all these egocentric benchmarks with only conventional VQA tasks neglect that the designed task format should be grounded in the potential applications.
Therefore, in this paper, we focus on comprehensively exploring the capabilities for different functions of MLLMs in Embodied AI. 
A comparison to recent video-language benchmarks is presented in Table~\ref{tab:comparison}.

\noindent\textbf{Egocentric Video Datasets.}
Egocentric video datasets~\citep{grauman2022ego4d, damen2018scaling, 6248010, sigurdsson2018charadesegolargescaledatasetpaired} capture first-person interactions with environment, aiding robotic tasks and augmented reality.
These datasets are often recorded via head-mounted cameras or wearable glasses.
% Some of them are further annotated at key frames with natural language descriptions.
As more egocentric videos become available, specialized datasets focusing on specific aspects of ego-perspective have emerged. 
For instance, LEMMA~\citep{jia2020lemma} includes data on goal-directed actions and multi-task situations.
Ego-ExoLearn~\citep{huang2024egoexolearn} and Ego-Exo4D~\citep{grauman2024ego} emphasize egocentric videos that demonstrate an individual's understanding of activities when given an exocentric demonstration.
These datasets provide a robust foundation for training and evaluating MLLMs from a first-person perspective.

\section{Task Types in \textit{VidEgoThink}}

Given that the utilization of foundation models in Embodied AI remains an open research question, we carefully design four types of interrelated tasks for comprehensive assessment as shown in Figure~\ref{fig:intro}: (i) \textit{video question-answering}~\citep{Cheng_2024_CVPR, fan2019egovqa, jia2022egotaskqa}, (ii) \textit{hierarchy planning}~\citep{ahn2022can, huang2022inner, huang2022language}, (iii) \textit{visual grounding}~\citep{gao2023physically, chiang2024mobility, munasinghe2023pg}, and (iv) \textit{reward modeling}~\citep{kwon2023reward, di2023towards, yu2023language}.
The detailed descriptions of four different tasks are as follows.

\subsection{Video Question Answering}

Previous evaluation studies on egocentric vision~\citep{Cheng_2024_CVPR} have predominantly focused on static images, constrained by the input format limitations of earlier MLLMs.
However, recent advancements in API-based and video-based MLLMs~\citep{achiam2023gpt, anthropic2024claude, reid2024gemini, li2023videochat, lin2023video} have demonstrated significant progress. 
Since our real world is inherently dynamic and humans frequently process substantial amounts of video data, it is crucial to evaluate the video understanding capabilities of MLLMs.

\begin{figure}[h!]
    \centering
    \includegraphics[width=\linewidth]{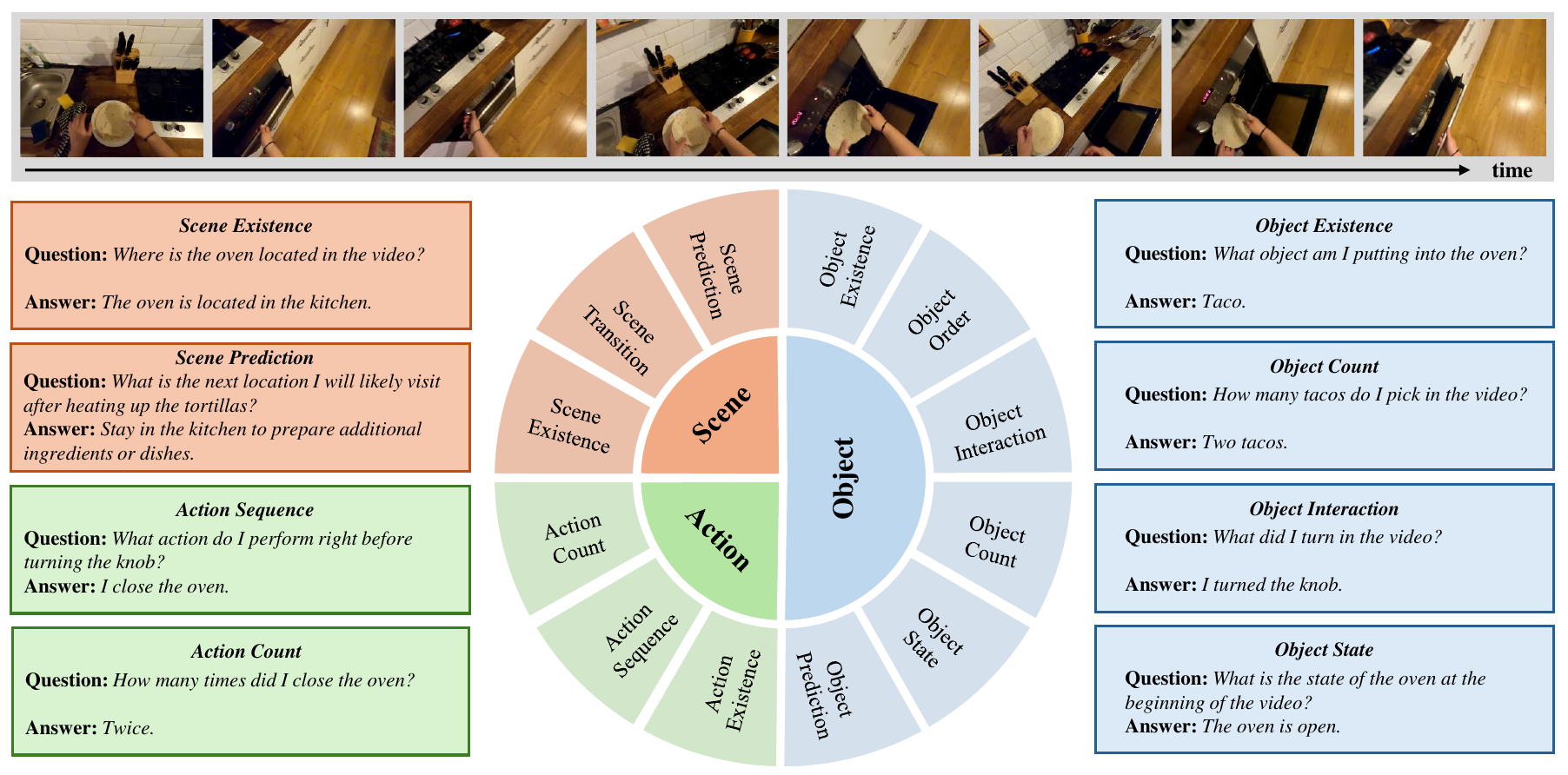}
    \caption{Case of video question answering.}
    \label{fig:case-vqa}
\end{figure}

\paragraph{Dimensions.}
To underscore the differences between static images and dynamic videos~\citep{li2023mvbench}, we emphasize temporal attributes, ensuring that questions require the entire video for accurate answers rather than just a single frame.
Considering the essential abilities for observing and interacting with the real world from a first-person perspective, we decompose the content of video modalities around ``myself'' into three main elements: object, action, and scene.
Furthermore, we explore a series of fine-grained dimensions from these elements as shown in Figure~\ref{fig:case-vqa}.

% Following~\citep{Cheng_2024_CVPR}, we explore five categories with eighteen fine-grained dimensions, reflecting the aspects humans typically consider when observing and interacting with the real world.
% Since planning is important and has become a standard task in Embodied AI, we will elaborate on it in Section~\ref{sec:planning}.

% By referencing the most critical abilities when interacting with the real world from a first-person perspective, we decompose the content of video modalities into three main entities: objects, actions, and environment. Furthermore, following the logical sequence of cognition, reasoning, and prediction, we have specifically designed five categories and 18 fine-grained dimensions to quantitatively assess capabilities from the first-person perspective.

\begin{itemize}[leftmargin=*]
\item \textbf{Object.}
Observing and interacting with objects are fundamental capabilities of the human visual system. 
Egocentric videos emphasize the objects seen or used by ``myself''.
We further divide the object category into six fine-grained dimensions: 
(1) \textit{Object Existence (OE)}: Determining whether an object has appeared in the video;
(2) \textit{Object Order (OO)}: Identifying the sequence of objects appearing in the video;
(3) \textit{Object Interaction (OI)}: Assessing whether and how an object has been used in the video;
(4) \textit{Object Count (OC)}: Counting the total number of objects for a specific type;
(5) \textit{Object State (OS)}: Assessing whether the state of an object has changed;
(6) \textit{Object Prediction (OP)}: Predicting what will happen to a certain object.

\item \textbf{Action.}
Action recognition involves automatically identifying specific human actions within the video.
Egocentric videos emphasize events that involve interactions with ``myself''.
Since action prediction is important and has become a standard task in Embodied AI, we will elaborate on it in Section~\ref{sec:planning}.
We further divide the action dimension into three fine-grained dimensions:
(1) \textit{Action Existence (AE)}: Determining whether an action has happened in the video;
(2) \textit{Action Sequence (AS)}: Identifying the sequence in which actions occur in the video;
(3) \textit{Action Count (AC)}: Counting the frequency with which an action occurs.

\item \textbf{Scene.}
Perceiving the surrounding scene from a first-person perspective is essential for interacting with the environment.
In egocentric video data, the constant movements of my body and viewpoint make describing an object's position relative to one's orientation challenging, necessitating integration with the environmental context.
We have also designed three fine-grained dimensions to perceive the scene:
(1) \textit{Scene Existence (SE)}: determining whether the video takes place in a certain scene;
(2) \textit{Scene Transition (ST)}: Identifying the transitions between scenes visited;
(3) \textit{Scene Prediction (SP)}: Predicting where the next scene will take place.

\end{itemize}

\paragraph{Task Format.}
Two mainstream methods for video question-answering include \textit{multiple-choice} and \textit{open-ended} question-answering.
Compared to multiple-choice, open-ended text generation is more natural in real-world applications.
Moreover, it is difficult to design distractors in multiple-choice question-answering to ensure that there are no inherent shortcuts.
Therefore, we mainly adopt open-ended text generation as our task format for traditional video question-answering.
% Recognizing the advantages and limitations of each format, we employ both to complement each other, thus providing a comprehensive evaluation kit.

\begin{itemize}[leftmargin=*]
% \item \textbf{Multiple-Choice Question-Answering.}
% Given an egocentric video $i$ containing the observation and interaction information from the first-person perspective, along with a question $q_i$ and five answer candidates $\{a_i^1, \cdots, a_i^5\}$ with the ground-truth answer $a_i^{gt}$, the model is asked to select the most suitable option $a_i$.
% Multiple-choice questions offer ease of evaluation, providing a straightforward accuracy metric based on close-set choices. 
% However, it is crucial to carefully design distractors to ensure that there are no inherent shortcuts.

\item \textbf{Open-Ended Question-Answering.}
Given an egocentric video $i$ along with a question $q_i$, the model is asked to generate responses $r_i$ in free-text form, akin to human communication. The generate answer $r_i$ is then compared to its corresponding ground-truth response $r_i^{gt}$.
% Furthermore, recent studies~\citep{zheng2024judging} using API-based LLMs as automatic evaluators can be costly in terms of both time and money, and the scores can be unstable with updates to the API-based LLM.
\end{itemize}

\paragraph{Metrics.}
% For both tasks, we use accuracy as the uniform metric.
% Accuracy in multiple-choice question-answering (\textit{Acc-MC}) can be directly computed by determining whether the selected option matches the ground truth.
Traditional metrics~\citep{chen2019evaluating, papineni-etal-2002-bleu} fall short in accurately assessing semantic similarity.
Follwing~\citet{zheng2024judging}, we employ API-based LLMs as automatic evaluators (\textit{Acc-VQA}) to assess the performance of generated answers.
These evaluators have demonstrated high correlations with human labels~\citep{zheng2024judging, Cheng_2024_CVPR}, making them a reliable substitute for human assessment.
% For open-ended question-answering, we employ both API-based and our specially fine-tuned LLMs as automatic evaluators to assess the performance of generated answers.
% The final accuracy is then computed as \textit{Acc-API} and \textit{Acc-FT}, respectively.

\begin{itemize}[leftmargin=*]
% \item \textbf{Acc-MC.}
% To compute model accuracy for the entire multiple-choice question-answering benchmark $\mathcal{D}$, we use the boolean function $\mathbb{I}(\cdot)$ that returns one when model output $\hat{a}_i$ equals the ground truth $a_i$ and zero otherwise.
% \begin{equation}
%     \displaystyle  \text{Acc-MC}=\frac{1}{|\mathcal{D|}}\sum_{i=1}^{|\mathcal{D}|}\mathbb{I}(\hat{a}_i=a_i)
% \end{equation}

\item \textbf{Acc-VQA.} 
Given the limitations of traditional metrics, we use API-based LLMs $g(\cdot)$ with superior reasoning abilities to evaluate open-ended answers.
Specifically, we assign the score $g(\hat{r}_i, r_i)$ as 0 (wrong), 0.5 (partially correct), or 1 (correct) to the generated response $\hat{r}_i$ with reference to the question $q_i$ and the corresponding ground-truth response $r_i$.
The performance of benchmark $\mathcal{D}$ is then computed by averaging all scores as follows:
\begin{equation}
    \displaystyle  \text{Acc-VQA}=\frac{1}{|\mathcal{D|}}\sum_{i=1}^{|\mathcal{D}|}g(\hat{r}_i, r_i),\; g(\hat{r}_i, r_i)= \begin{cases} 1 & \text{correct} \\ 0.5 & \text{partially correct} \\ 0 & \text{incorrect} \end{cases}
\end{equation}

% These LLMs are capable of handling uncertainties in answer generation, capturing the most important information. 
% For multiple-choice questions, most models are capable of producing succinct responses; however, due to the disparities between language capabilities and the free-form nature of MLLM output, a conclusive answer may not always be extracted using traditional natural language methods.
% Here LLMs serve as the best method to extract model responses. For open-ended questions, LLM pays close attention to semantics and is best at comparing reference responses to the generated output. 

% \item \textbf{Acc-FT.}
% \hi{@Sicheng}

% Despite the promising performance of API-based LLMs, their usage comes with issues.
% API-based models can be costly for large-scale benchmarks, and updates to these models can introduce inconsistencies in reported scores over time, undermining benchmark stability.
% To address these issues, we propose fine-tuning an open-source LLM as our custom evaluator by leveraging evaluation results from API-based evaluators.
% Our custom evaluator, trained to achieve high human correlation and test accuracies, computes scores similarly to API-based LLMs.
% This approach effectively negates issues with API-based model where the custom evaluator will incur no monetary costs and maintain consistency through time. 
\end{itemize}

\subsection{Hierarchy Planning}
\label{sec:planning}

Recently, a hierarchy planning framework~\citep{ahn2022can, singh2023progprompt, vemprala2024chatgpt} has been proposed to combine the advantages of foundation models and traditional methods in Embodied AI.
In detail, foundation models are used as the planner to decompose high-level task instructions (e.g., ``\textit{cook salmon}'') into either mid-level steps (e.g., ``\# put salmon in the microwave') or low-level atomic actions (e.g., ``\texttt{find}(\textit{microwave})''), which is much more convenient for controlling.
% Therefore, because of the importance of planning, we treat it as a formal task alone.
Despite EgoPlan-Bench~\citep{chen2023egoplan} exploring the planning capability from the first-person perspective, it only considers decomposing the high-level goal into mid-level steps, and its task format is multi-choice which is less natural.

% \hi{@Sicheng: Please add a figure of cases here, referring to Figure 1 in Ego4d Goal-Step.}
\begin{figure}[h!]
    \centering
    \includegraphics[width=1\linewidth]{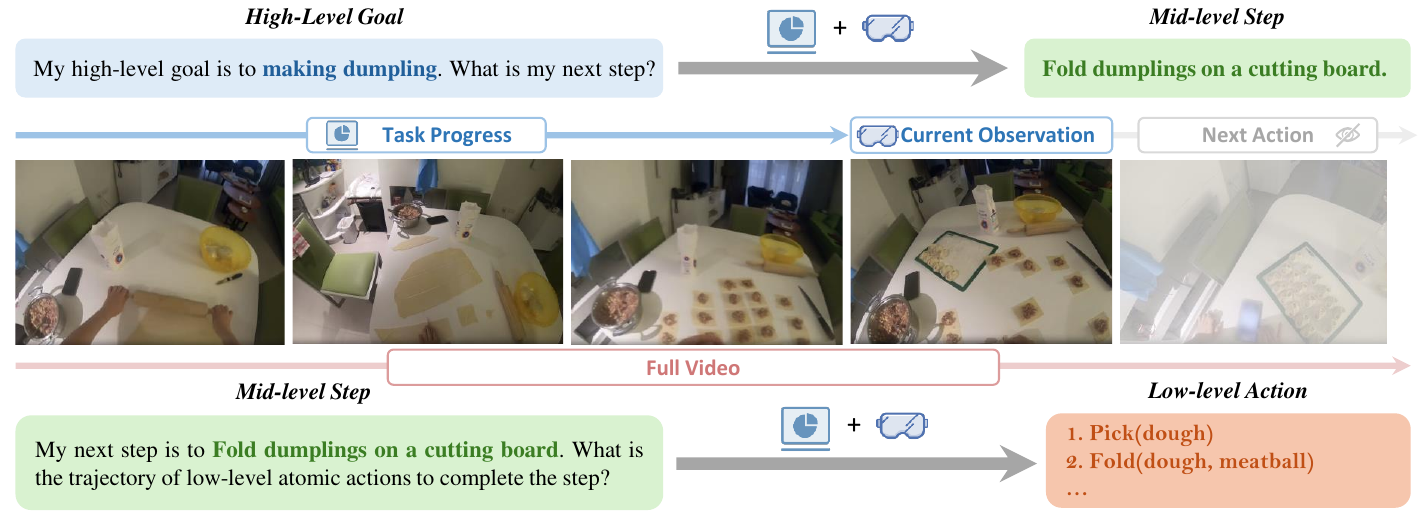}
    \caption{Case of hierarchy planning.}
    \label{fig:case-hp}
\end{figure}

\textbf{Task Format.} As illustrated in Figure~\ref{fig:case-hp}, we design two types of planning tasks: high-level goal to mid-level step (\textit{High-to-Mid}), and mid-level step to low-level action (\textit{Mid-to-Low}).

\begin{itemize}[leftmargin=*]
\item \textbf{High-to-Mid.}
% Given an egocentric video $i$ which includes history and current observations, a high-level goal $G_i$, MLLMs are required to generate the next step $\hat{s}_i$ in the form of free-text, compared to the ground-truth step $s_i$ appeared after the provided video.
% Rather than directly generating the whole long-term planning, we believe that our designed task format can better leverage the information of each step in videos, and involve the future feedback signals to improve.
Given an egocentric video $i$ with historical and current observations, a high-level goal $G_i$, MLLMs are required to generate the next step $\hat{s}_i$ in free-text format. This generated step is then compared to the ground-truth step $s_i$ that follows the provided video.
We adopt a step-by-step format rather than directly generating the entire long-term plan because our focus is on evaluation rather than method development.

\item \textbf{Mid-to-Low.}
% Given the pre-defined set of low-level atomic actions $\mathcal{A}$ that encompasses the common functions in daily human activities, an egocentric video $V_i$, the ground-truth of a mid-level step $s_i$ that is yet to be complete, MLLMs are required to generate the trajectory of low-level actions $\hat{\mathcal{T}}_i = (\hat{a}_1, \cdots, \hat{a}_n)$ using functions from $\mathcal{A}$ to complete the mid-level step. The corresponding ground-truth trajectory of actions that appeared after the provided video is $\mathcal{T}_i = (a_1, \cdots, a_m)$.
Given a pre-defined set of low-level atomic actions $\mathcal{A}$ that encompasses common functions in daily human activities, an egocentric video $V_i$, and the ground-truth of a mid-level step $s_i$ that is yet to be complete, MLLMs are required to generate the trajectory of low-level actions $\hat{\mathcal{T}}_i = (\hat{a}_1, \cdots, \hat{a}_n)$ using functions from $\mathcal{A}$ to complete the mid-level step. The corresponding ground-truth trajectory of actions that appeared after the provided video is $\mathcal{T}_i = (a_1, \cdots, a_m)$.
\end{itemize}

\textbf{Metrics.}
Considering the difficulty of hierarchical planning tasks, we directly use API-based LLMs to compute accuracy (\textit{Acc-H2M} and \textit{Acc-M2L}).
However, these metrics are a trade-off due to the challenges of evaluation video planning tasks. We will introduce an advanced version in future work, as discussed in Sec.~\ref{sec:limitation}.

\begin{itemize}[leftmargin=*]
    \item \textbf{Acc-H2M.} In the high-to-mid task, we apply API-based LLMs $g(\cdot)$ to compute the similarity score $g(\hat{s}_i, s_i)$ between generated step prediction $\hat{s}_i$ and ground truth $s_i$ for the benchmark $\mathcal{D}$. Specifically, we also assign the score $s_i$ as 0 (wrong), 0.5 (partially correct), or 1 (correct).
    
    \begin{equation}
        \displaystyle \text{Acc-H2M} = \frac{1}{|\mathcal{D}|}\sum_{i=1}^{|\mathcal{D}|}g(\hat{s}_i, s_i)\;,\;\;g(\hat{s}_i, s_i) = \begin{cases}
             1 & \text{correct} \\ 0.5 & \text{partially correct} \\ 0 & \text{incorrect}
        \end{cases}
    \end{equation}
    % \item \textbf{Exec.} For each mid-to-low task in the benchmark $\mathcal{D}$, we want to assess the executability of the generated low-level actions. We compute this by comparing whether the generated atomic actions $\hat{\mathcal{T}}_i$ appear in the same order as ground-truth atomic actions $\mathcal{T}_i$ while disregarding potential intermediate steps outputted from the model. \csj{todo} We denote $\mathcal{T}\cap\mathcal{\hat{T}}$ as an ordered set of function calls that appeared both in the model output and the ground truth.
    % \begin{equation}
    %     \displaystyle \text{Exec} = \frac{1}{n} \sum_{a_i \in \mathcal{T},\;\hat{a}_i \in \mathcal{T}\cap\mathcal{\hat{T}}} \mathbb{I}(\hat{a}_i=a_i)
    % \end{equation}
    % \item \textbf{mIoU-M2L.} We disregard the order of low-level actions and solely consider whether the actions in the ground truth ($\mathcal{T}$) appear in the model output ($\mathcal{\hat{T}}$). We compute the metric by dividing the cardinality of the intersection of ground truth actions intersection and generated actions by the cardinality of their union. This value is then averaged across the entire set of low-level benchmarks. 
    % \begin{equation}
    %     \displaystyle \text{mIoU-M2L} = \frac{1}{|\mathcal{D}|}\sum_{i=1}^{|\mathcal{D}|}\frac{|\mathcal{\hat{T}}_i\cap\mathcal{T}_i|}{|\mathcal{\hat{T}}_i\cup\mathcal{T}_i|}
    % \end{equation}
    \item \textbf{Acc-M2L.} For the Mid-to-Low task, which is akin to tool learning~\citep{guo2024stabletoolbench, qin2023toolllm} by calling low-level functions and evaluating the success rate, we also use API-based LLMs to determine the completion status. We assign the score $g(\mathcal{\hat{T}}, \mathcal{T})$ to compute the similarity between the generate action trajectory $\mathcal{\hat{T}}$ and the ground-truth trajectory $\mathcal{T}$, using a scale from 0 to 10 to increase the degree of differentiation. 

    \begin{equation}
        \displaystyle \text{Acc-M2L} = \frac{1}{|\mathcal{D}|}\sum_{i=1}^{|\mathcal{D}|}g(\mathcal{\hat{T}}, \mathcal{T})\;,\;\; 0 \leq g(\mathcal{\hat{T}}, \mathcal{T}) \leq 10
    \end{equation} 
\end{itemize}

\subsection{Visual Grounding}
While natural language is effective for human communication, it cannot be directly translated into low-level actions or grounded in the real world.
Consequently, visual grounding~\citep{peng2023kosmos, chen2023shikra, munasinghe2023pg} has garnered significant attention in both image- and video-based MLLMs.
This task requires models to ground complex natural language descriptions or instructions in an image or video and output the corresponding pixel-level bounding boxes, masks, or frames.
The bounding boxes and masks can directly identify actionable objects~\citep{munasinghe2023pg, zheng2024instruction}, while the frames can provide sufficient spatial or temporal information for downstream tasks~\citep{li2024lego, chiang2024mobility}.

% The task is set as given an image and a natural language description, and the model needs to give a pixel-level bounding box or mask.
% Recently, with the rise of Large Language Models and Vision Language Models, many grounded models based on LLMs have emerged and made significant progress such as KOSMOS-2~\citep{peng2023kosmos}, Shikra~\citep{chen2023shikra} and LLaVA-Grounding~\citep{zhang2023llava}, etc.
% Similarly, in the video field, many video language models that support grounding have emerged, such as PG-Video-LLaVA~\citep{munasinghe2023pg} and VITRON~\citep{fei2024vitron}, etc. When grounding a video, in addition to visual information, we also have temporal information.
% We can combine the traditional object tracking task and the visual grounding task into one task, which requires a video clip and a natural language description, and then outputs the bounding box of the object related to the description in each frame of the video.

% As mentioned in RefEgo~\citep{Kurita_2023_ICCV}, for first-person videos, some frames may not contain the objects we need to refer to. Therefore, being able to determine which fragments are useful is an important ability, which is temporal grounding. This task is supported in LEGO~\citep{li2024lego}, ChatVTG~\citep{qu2024chatvtg} and TimeChat~\citep{ren2024timechat}, etc., which requires an input of a video and a natural language description, and the model needs to give time segments related to the description.
% \paragraph{Definition.}

\begin{figure}[h!]
    \centering
    \includegraphics[width=1\textwidth]{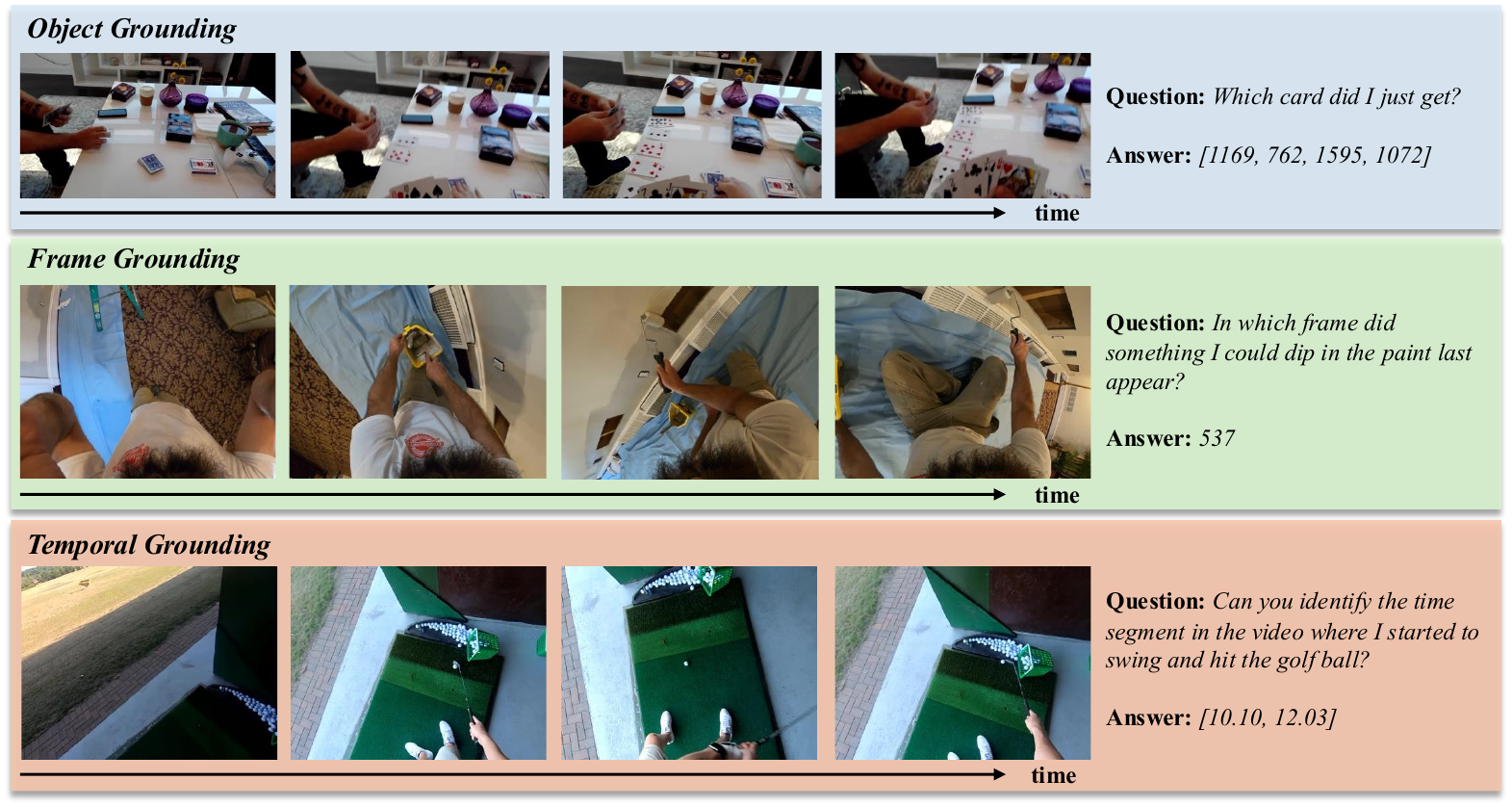}
    \caption{Cases of visual grounding.}
    \label{fig:case-vg}
\end{figure}

% \begin{figure}[h!]
%     % \centering
%     % \includegraphics[width=0.5\linewidth]{images/grounding-case1.png}
%     % \caption{Examples of visual grounding task in VidEgoThink.}
%     % \label{fig:enter-label}
    
%     \centering
%     \begin{minipage}[b]{0.3\textwidth}
%         \includegraphics[width=\textwidth]{images/grounding-case1.png}
%         \caption{\textbf{Case for Object Grounding:} Which was the last card I played?}
%         \label{fig:image1}
%     \end{minipage}
%     \hfill % 用于两图之间的间隔
%     \begin{minipage}[b]{0.3\textwidth}
%         \includegraphics[width=\textwidth]{images/grounding-case2.png}
%         \caption{\textbf{Case for Frame Grounding:} In which frame did something I could dip in the paint last appear?}
%         \label{fig:image2}
%     \end{minipage}
%     \hfill
%     \begin{minipage}[b]{0.3\textwidth}
%         \includegraphics[width=\textwidth]{images/grounding-case3.png}
%         \caption{\textbf{Case for Temporal Grounding:} What are the all intervals for strikes?}
%         \label{fig:image3}
%     \end{minipage}
% \end{figure}

\paragraph{Task Format.}
Although RefEgo~\citep{Kurita_2023_ICCV} considers object tracking from the first-person perspective, its output format is designed for conventional computer vision methods rather than MLLMs.
To bridge this gap, we specifically design three tasks for different situations as shown in Figure~\ref{fig:case-vg}: \textit{object grounding}, \textit{frame grounding}, and \textit{temporal grounding}.

\begin{itemize}[leftmargin=*]
\item \textbf{Object Grounding.} 
Given an egocentric video $i$ and a natural language query $q_i$ of an object, the model is required to provide a bounding box represented as a rectangle $B_i=\left[x_1,x_2,y_1,y_2\right]$ containing the query object in the last frame of the video. 
Performance is evaluated by comparing with the ground truth $B_i^{gt}=\left[x_1^{gt},x_2^{gt},y_1^{gt},y_2^{gt}\right]$.
It is worth noting that the query $q_i$ is designed according to the whole video rather than only the last frame.
Locating target objects that appeared earlier in the observed frame is crucial for downstream tasks, like manipulation and navigation.
% or a mask which is represented as a binary image $(I_i)$ of the same size as the original image 
% Object grounding is a traditional computer vision task that requires the model's cross-modal fusion capability.
% When an agent is in a real-world scenario and needs to interact with the surrounding environment, it needs to locate the objects to be operated before the agent uses tools such as robotic arms to interact.
% When the input changes to the video, the output bounding box needs to be kept on the object of interest because the agent needs the latest information about the object in order to adjust the action of the agent at any time.

\item \textbf{Frame Grounding.}
Given an egocentric video $i$ and a natural language query $q_i$, the model is asked to identify the keyframe $K_i$ containing the required information, which will be compared with the set of ground-truth keyframes $\{K_{ij}^{gt}\}$ in a small enough time interval of the last appearance of the target. That is, if an object or action occurs multiple times, we specify the time interval of the last appearence time as the ground truth, as it generally contains the most useful information for the current situation. 
In embodied scenes, there is always a need to retrieve objects, people, or events that occurred earlier.

\item \textbf{Temporal Grounding.} Given an egocentric video $i$ and a natural language query $q_i$, the model is required to identify the time segments in the video corresponding to the query, represented as $T_i=\left[l_{i},r_{i}\right]$, where $0\leq l_i\leq r_i\leq \vert V_i\vert$, and $\vert V_i\vert$ is the total number of frames in the video.
The ground truth $T_i^{gt}$ follows a consistent format.
It is crucial to know the relevant time segment that occurred previously, such as the frequency of an event and the complete trajectory of an object.
% When an agent needs to perform a detailed analysis of what has happened or the trajectory of an object, it must know all the relevant time segments before. 
% For example, when the agent needs to know how many spoonfuls of salt the user in front of it has added to the dish and then issues a reminder, it needs to obtain each time the user adds salt, which is reflected in the input as several related time segments. (just one possible example)
\end{itemize}

\paragraph{Metrics.}
For object grounding and temporal grounding, we use mean intersection over union (\textit{mIoU}) as the uniform metric, named \textit{mIoU-Object} and \textit{mIoU-Temporal}, respectively.
They calculate the similarity between the output and the ground truth because their output results can be expressed as a certain region or a certain range.
% For the outputs of mask type in object grounding, we should first convert it into a bounding box. 
For frame grounding, we use accuracy (\textit{Acc-Frame}) to evaluate the model output as the answer is an integer.

\begin{itemize}[leftmargin=*]
\item \textbf{mIoU-Object.}
We denote the bounding box output as $\hat{B}_i$ and the ground truth as $B_i$, where $i$ represents a sample. 
The similarity in the benchmark $\mathcal{D}$ is calculated using mIoU as follows.
\begin{equation}
    \text{mIoU-Obj}=\frac{1}{\vert\mathcal{D}\vert }\sum_{i=1}^{|\mathcal{D}|}\frac{\vert \hat{B}_{i}\cap B_{i}\vert}{\vert \hat{B}_{i}\cup B_{i}\vert}
    \label{eq:miou}
\end{equation}

\item \textbf{Acc-Frame.} 
Given the keyframe index $\hat{k}_i$ produced by the model and its corresponding ground truth set $\mathcal{K}_i$, we can calculate the accuracy in the benchmark dataset $\mathcal{D}$ as follows. 
Here, $\chi(\cdot)$ is an indicator function that equals 1 if $\hat{k}_i$ in $\mathcal{K}_i$, and 0 otherwise.
\begin{equation}
    % \text{MSE}=\frac{1}{\mathcal{\vert D\vert}}\sum_{i=1}^{|\mathcal{D}|} ( K_i-K_i^{gt})^2
    \text{Acc-Frame}=\frac{1}{\mathcal{\vert D\vert}}\sum_{i=1}^{|\mathcal{D}|} \chi_{\mathcal{K}_i}\left(\hat{k}_i\right)
\end{equation}

\item \textbf{mIoU-Temporal.}
We denote the time interval covered by the model output as $\hat{T}_i$ and ground truth as $T_i$ respectively, where $i$ represents a video sample. Similarly, we can calculate the score in the benchmark $\mathcal{D}$ as follows. 
\begin{equation}
    \text{mIoU-Temp}=\frac{1}{\vert\mathcal{D}\vert }\sum_{i=1}^{|\mathcal{D}|}\frac{\vert \hat{T}_{i}\cap T_{i}\vert}{\vert \hat{T}_{i}\cup T_{i}\vert}
\end{equation}
    
\end{itemize}

\subsection{Reward Modeling}

In Embodied AI, manually designing reward functions to supervise actions is challenging due to the need for accuracy and diversity, especially for human activities.
Benefiting from the large-scale Internet training corpus, foundation models can serve as reward models with built-in commonsense and reasoning capabilities.
There are three primary approaches to deploying foundation models as reward models:
(1) Using a sparse proxy reward function with a simple binary score~\citep{kwon2023reward};
(2) Computing similarity between close-vocabulary action phrases and images~\citep{di2023towards, rocamonde2023vision};
(3) Generating code to translate task semantics into combined reward functions~\citep{yu2023language, ma2023eureka}.
Considering the feasibility of targeting video data, this paper primarily focuses on the first approach.

\begin{figure}[h!]
    \centering
    \includegraphics[width=1\linewidth]{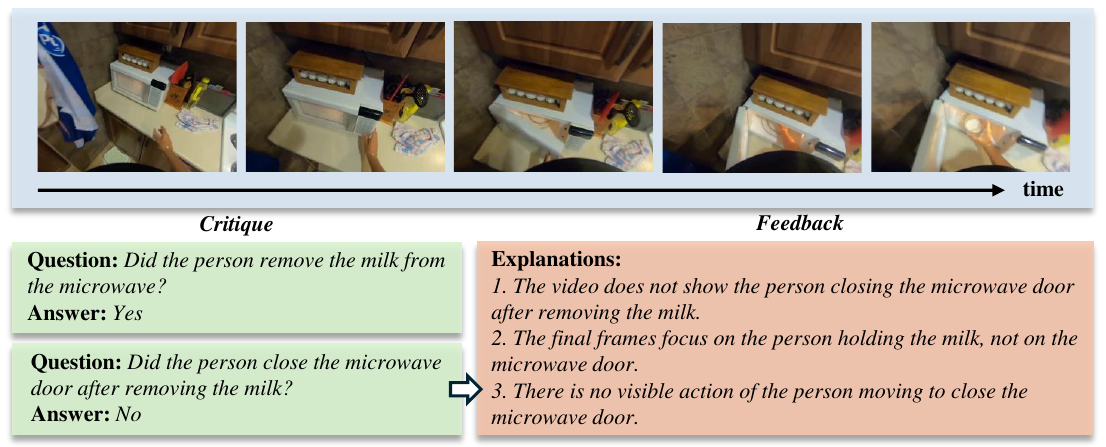}
    \caption{Case of reward modeling.}
    \label{fig:case-rm}
\end{figure}

\paragraph{Task Format.}
As a reward model, MLLMs should first observe the video to determine the completion status of the target motion. If the action is not completed, the reward model should further provide fine-grained feedback to help achieve the goal inspired by~\citep{wang2020semeval, cheng2023unsupervised}.
Hence, we specifically design two types of tasks like Figure~\ref{fig:case-rm}: \textit{critique}, and \textit{feedback}.

\begin{itemize}[leftmargin=*]
\item \textbf{Critique}. Given an action-specific egocentric video $i$ and its corresponding natural language action description $a_i$, the reward model needs to directly generate a binary answer $\hat{y}_i$ (i.e., \texttt{Yes} or \texttt{No}) that indicates whether the action has been completed in the observed video.

\item \textbf{Feedback}.
% Given the natural language description of an incomplete task ($T_i$) and its corresponding video segments ($V_i$), MLLMs are required to provide feedback ($\hat{r}_i$) on how $T_i$ can be completed based on $V_i$.
Given an uncompleted action $a_i$ in the $i$-th egocentric video, the reward model needs to provide fine-grained feedback $\hat{r}_i$ on why the action is not completed based on current observations, compared to the ground-truth references $\mathcal{R}_i = \{r_i^1, r_i^2, r_i^3\}$. 
This feedback is valuable for guiding and correcting downstream models to learn the policy for completing the target action.
\end{itemize}

\paragraph{Metrics.}
We apply the following accuracy metrics to assess performance in critique and feedback tasks (\textit{Acc-Critique} and \textit{Acc-Feedback}) for the reward modeling tasks.

\begin{itemize}[leftmargin=*]
    \item \textbf{Acc-Critique}. For each critique task in the benchmark $\mathcal{D}$, we compare the generated critique $\hat{y}_i$ with its ground-truth label $y_i$. The boolean function $\mathbb{I}(\cdot)$ returns one for each identical output label and zero otherwise. 
    % We sum over this function across all critique tasks in $\mathcal{D}$ and normalize by the cardinality to get model accuracy.
    \begin{equation}
        \displaystyle \text{Acc-Cri} = \frac{1}{|\mathcal{D}|}\sum_{i=1}^{|\mathcal{D}|}\mathbb{I}(\hat{y}_i=y_i)
    \end{equation}
    
    \item \textbf{Acc-Feedback}. To quantitatively assess the similarity between the generated feedback $\hat{r}_i$ and the set of reference feedback $\mathcal{R}_i$, we employ evaluator LLM $g(\cdot)$ that assigns a score with 0 (wrong), 0.5 (partially correct) or 1 (correct).
    \begin{equation}
        \displaystyle \text{Acc-Feed} = \frac{1}{|\mathcal{D}|}\sum_{i=1}^{|\mathcal{D}|}g(\hat{r}_i,\mathcal{R}_i),\;g(\hat{r}_i,\mathcal{R}_i)=\begin{cases}
            1 & \text{correct} \\
            0.5 & \text{partially correct} \\
            0 & \text{wrong}
        \end{cases}
    \end{equation}
\end{itemize}

\section{Data Collection in \textit{VidEgoThink}}

Recent releases of egocentric video datasets~\citep{grauman2022ego4d, grauman2024ego, huang2024egoexolearn} have advanced the field of Embodied AI.
To ensure diversity and popularity, we use the popular Ego4D dataset~\citep{grauman2022ego4d} to construct VidEgoThink benchmark. 
Ego4D-v2\footnote[1]{https://ego4d-data.org/docs/updates/} contains 3,900 hours of 9,611 egocentric videos with diverse human annotations.
To avoid data leakage, we select videos from the validation dataset.
However, due to the video length limitations of MLLMs, the lengthy Ego4D videos, ranging from tens of minutes to over an hour, are unsuitable.
Additionally, manually labeling question-answering data requires significant human effort. 
To address these problems, we design strategies to automatically clip the videos to appropriate lengths and generate corresponding question-answer pairs.
To prevent the VidEgoThink benchmark from being compromised through prompt engineering, the detailed prompts used for automatic annotation construction will not be released.
The statistics of each task in VidegoThink are presented in Table~\ref{tab:sta_data}.

% \subsection{Automatic Annotation} 

% Manually labeling question-answering data for various tasks requires significant human effort. 
% To address this, we have designed strategies for automatic annotation using the diverse existing annotations in Ego4D and the advanced commonsense capabilities of GPT-4o~\citep{gpt4o}.
% To prevent the VidEgoThink benchmark from being hacked through prompt engineering, the detailed prompts used for automatic annotation construction will not be released.

\begin{table}[h!]
    \centering
    \caption{The statistics of videos across different benchmarks. Duration denotes the average time duration in second of all videos. LenQ and LenA indicate that the average length of questions and answers in the word level. TypeQ denotes the type of questions.}
    \resizebox{\textwidth}{!}{
    \begin{tabular}{llccccccccc}
    \toprule
        \multirow{2}{*}{\textbf{Benchmark}} & \multirow{2}{*}{\textbf{Subtask}} & \multicolumn{3}{c}{\textbf{Video}} & \multicolumn{4}{c}{\textbf{Question-Answering}} & \multirow{2}{*}{\textbf{\#Scene}}\\
        \cmidrule(rl){3-5}
        \cmidrule(rl){6-9}
         & & \#Original & \#Clipped & Duration & \#Instance & LenQ & LenA & TypeQ \\ 
        \midrule
        % \rowcolor[gray]{0.95} \multicolumn{8}{c}{\textbf{Ego4D}}\\
        % \midrule
        % \textbf{Goal-Step} (Ego4D) & Validation & 134 & 3,706 & - & - & - & - & - & 12 \\
        % \midrule
        % \specialrule{0em}{0.5pt}{0.5pt}
        % \midrule
        \multirow{3}{*}{\textbf{Video Question Answering}} & Object & 29 & 57 & 23.71 & 300 & 10.88 & 7.13 & 5 & 9 \\
         & Action & 39 & 78 & 24.56 & 150 & 10.85 & 4.72 & 4 & 9 \\
         & Scene & 45 & 82 & 21.91 & 150 & 11.46 & 8.34 & 4 & 9 \\
        \midrule
        \multirow{2}{*}{\textbf{Hierarchy Planning}} & High-to-Mid & 76 & 598 & 1008.26 & 598 & 16.5 & 5.18 & 1 & 9 \\
         & Mid-to-Low & 76 & 598 & 1008.26 & 598 & 22.12 & 6.02 & 1 & 9 \\
        \midrule
        \multirow{3}{*}{\textbf{Visual Grounding}} & Object & 41 & 88 & 119.05 & 220 & 22.60 & - & 1 & 25 \\
         & Frame & 65 & 147 & 139.57 & 368 & 23.01 & - & 1 & 25\\
         & Temporal & 69 & 416 & 68.90 & 735 & 82.40 & - & 1 & 8 \\
        \midrule
        \multirow{2}{*}{\textbf{Reward Modeling}} & Critique & 76 & 963 & 16.60 & 1236 & 11.21 & {\color{white} 1}1.00 & 1 & 9 \\
         & Feedback & 74 & 638 & 15.08 & 638 & 19.24 & 53.06 & 1 & 9  \\
        % \midrule
        % \textbf{Total} (VidEgoThink) & - & \\
        \toprule
    \end{tabular}}
    \label{tab:sta_data}
\end{table}

\noindent\textbf{Video Question-Answering.}
To construct this benchmark, we integrate \textit{Narration} data, capturing interactions between the camera wearer and the environment, focusing on \textit{object}, \textit{action}, and \textit{scene}.
% , which records every action of the camera wearer in our clipped videos.
We develop specific prompts, combined with the narrations, as inputs for GPT-4o tailored to each fine-grained dimension.
GPT-4o then generate diverse question-answering pairs for these dimensions. 
Due to the noise in generated instances and the cost of API-based evaluation, three human annotators filter them to ensure quality and diversity, selecting the most representative examples.
Finally, we totally collect 600 instances with 50 examples per fine-grained dimensions.

\noindent\textbf{Hierarchy Planning.}
We use existing human annotations in Ego4D with goals-steps-substeps labels to construct our hierarchical data.
For video inputs, we use from 00:00 to the start time of the current step for both high-to-mid and mid-to-low subtasks.
In the high-to-mid task, high-level goals serve as inputs and corresponding mid-level step as labels.
Steps requiring numerous low-level actions and exceeding 180 seconds are decomposed into essential substeps.
Next, we use the ground-truth mid-level step and its \textit{Narration} as potential low-level atomic actions. 
To align with Embodied AI controller, GPT-4o converts narrations (e.g., ``\textit{C cuts a mango with a knife}'') into function calls (e.g., ``\texttt{cut}(\textit{mango}, \textit{knife})'') and merges semantically similar functions. 
% If a narration is too complex to be considered an atomic action, such as, ``\textit{C slices the mango into the container with the knife in her right hand}'', GPT-4o will decompose it into several appropriate functions like ``\texttt{slice}(\textit{mango}, \textit{knife})'' and ``\texttt{put}(\textit{sliced mango}, \textit{container})''.
% Moreover, we reserve examples with fewer than 12 low-level actions to reduce the difficulty of this task.
To ensure MLLMs understand the available low-level functions and their usage, we apply GPT-4o to generate their documentation. 
% Moreover, we manually merge functions and objects with similar semantics in low-level actions.
After filtering by three annotators, we obtain 598 clipped videos and instances for both tasks, with the mid-to-low task comprising 74 atomic actions.
% It is worth noting that the data for the other three tasks are also collected data from these 76 videos.

\noindent\textbf{Visual Grounding.}
\textit{Visual Queries} in Ego4D includes queries about objects and their tracks in the video, represented as frames with bounding boxes.
We use these annotations to collect object grounding and frame grounding subtasks.
For object grounding, given a clipped video and its annotations, we select the video from the beginning to the last annotated frame.
We construct a prompt with the \textit{Narration} in this segment for GPT-4o to generate a query.
The answer is the bounding box annotation of the object in the final frame.
In frame grounding, the video input spans from the start of the clipped video to either the ``query\_video\_frame'' annotated in \textit{Visual Queries} or the end frame of the clip.
We prompt GPT-4o using the object name and narrations within the time segment to generate a specific description of the frames containing the object.
All annotated frames in the input video are considered the answer.
Considering that step-substep annotations in \textit{Goal-Step} include temporal information, we primarily use these clipped videos.
By providing annotations and prompts to GPT-4o, we obtain a specific description of the selected sub-step as the query and the temporal interval of the sub-step as the answer. Finally, we obtain 369 instances for object and frame grounding, and 735 instances for temporal grounding.

\noindent\textbf{Reward Modeling.}
% To obtain a sufficient amount of egocentric video ($V_i$), we have chosen the Ego4D dataset~\citep{grauman2022ego4d} as our data source for the reward part.
Our clipped videos in the \textit{hierarchy planning} task contain entire mid-level steps, which we use to construct the reward modeling dataset.
% thereby avoiding the generation of meaningless video clips that lack sufficient action information.
We label the original complete videos as positive instances.
For negative instances, we employ two strategies: 
(1) using GPT-4o to generate questions where the action is similar but different from the video content; 
(2) manually crop each video clip to 60\%--80\% of its original length to ensure the action remains unfinished.
Each negative sample includes three feedback demonstrating the incomplete action.
% Reward models require extensive visual information to determine whether an action is complete; however, 
Considering narrations often lack detailed descriptions to determine whether an action is complete, we employ FFmpeg\footnote{https://www.ffmpeg.org/} to extract keyframes from each clipped video.
Then, we use GPT-4o to generate feedback from different aspects for negative instances based on step annotations and the extracted keyframes. After filtering by three annotators, we obtain 963 and 638 instances for critique and feedback tasks.

\section{Experiments}

In this section, we mainly introduce our extensive adopted models, including API-based models, a series of open-source image-based and video-based MLLMs.
The prompts for both inference and evaluation are shown in Appendix~\ref{app:prompt}.
Furthermore, we summarize the experimental results for different tasks, and their corresponding case studies are illustrated in Appendix~\ref{app:cases}.

\subsection{Models}

\begin{table}[h!]
    \centering
    \caption{LM, VM, TM, AM refer to the language module, visual module, temporal module, and alignment module. %GMHRA denotes the global multi-head relation aggregator. CLIP-ViT-O is the CLIP module pre-trained on OpenFlamingo, while 
    CLIP-ViT-L is CLIP module pre-trained on LLaVA, while CLIP-ViT-G is the CLIP module pre-trained on LAION. TA Frame Encoder denotes Time-aware Frame Encoder.}
    \resizebox{\textwidth}{!}{
    \begin{tabular}{lccccccccc}

        \toprule
        \multirow{2}{*}{\textbf{Model}} & \multirow{2}{*}{\textbf{LM}} & \multirow{2}{*}{\textbf{VM}} & \multirow{2}{*}{\textbf{TM}} & \multirow{2}{*}{\textbf{AM}} & \multirow{2}{*}{\textbf{Model Size}} & \multicolumn{2}{c}{\textbf{Training Data}} \\

         & & & & & & Image/Video-Text & Instruction  \\
        \midrule
        % \rowcolor[gray]{0.95} \multicolumn{8}{c}{\textbf{API-based Model}}\\
        % \midrule
        % \textbf{GPT-4-turbo} & \multicolumn{7}{c}{Unknown} \\
        % Gemini Pro Vision & \multicolumn{7}{c}{}\\
        % Tongyi Qianwen & \multicolumn{7}{c}{}\\
        % \textbf{GPT-4o} & \multicolumn{7}{c}{Unknown} \\
        % \textbf{GPT-4o-mini} & \multicolumn{7}{c}{Unknown} \\
        % \midrule
        \rowcolor[gray]{0.95} \multicolumn{8}{c}{\textbf{Image-based MLLMs}}\\
        \midrule
        \textbf{mPLUG-Owl2} & LLaMA & CLIP-ViT-L & - & Visual Abstractor & 7B & 1.23M & - \\
        \textbf{Qwen-VL} & Qwen & CLIP-ViT-G & - & VL Adapter & 7B & 1.4B & 350K\\
        %---------------------------------------
        \textbf{LLaVA-1.5} & LLaMA/Vicuna & CLIP-ViT-L-3 & -  & Linear & 7B & 558K & 665K \\
        % CHECKED
        %---------------------------------------
        \textbf{LLaMA-Adapter v2} & LLaMA & CLIP-ViT-L & - & Linear & 7B & 567K & 52K \\
        % CHECKED
        %---------------------------------------
        \midrule
        \rowcolor[gray]{0.95} \multicolumn{8}{c}{\textbf{Video-based MLLMs}}\\
        \midrule
        %---------------------------------------
        \textbf{LWM} & LLaMA2 & VQGAN & - & - & 7B & 1.01B & 519K \\
        \textbf{TimeChat} & LLaMA2 & CLIP-ViT-G & TA Frame Encoder & Sliding Video Q-Former & 7B & - & 177K  \\
        \textbf{GroundingGPT} & Vicuna-v1.5 & CLIP-ViT-L & Position Encoding & MLP & 7B & \textgreater1.3M & \textgreater770K\\
        \textbf{InternVL2} & InternLM2.5 & InternViT-300M & - & QLLaMA & 8B & 10B & - \\
        \textbf{InternLM-XComposer2.5} & InternLM2 & CLIP-ViT-L & - & Partial-LoRA & 7B & - & -  \\
        \textbf{Video-LLaVA} & Vicuna-v1.5 & Language Bind & - & Linear & 7B & 1.26M & 765K \\
        \textbf{PG-Video-LLaVA} & Vicuna-v1.5 & CLIP-ViT-L-3 & Grounding Module & MLP & 7B & - & 100K  \\
        \textbf{mPLUG-Owl3} & Qwen2 & SigLip-400M & MI-RoPE & Linear & 8B & \textgreater1.7M & \textgreater1M \\
        \textbf{MiniCPM-V2.6} & Qwen2 & SigLip-400M & - & Adaptive Visual Encoding & 8B & 570M & 3M \\
        \textbf{Qwen2-VL} & Qwen2 & ViT & M-RoPE & 3D-conv & 8B & 1.4T$_\text{tokens}$ & - \\
        % CHECKED
        %--------------------------------------
        \bottomrule
        \end{tabular}}
    \label{tab:sta_mllms}
\end{table}

\noindent\textbf{API-based Models.}
We conduct experiments with the representative GPT-4o (2024-05-13).
Since GPT-4o does not support video input, we address this limitation and enhance methodological diversity with the following assessment scheme: (1) \textit{w/ 32 frames}: Select 32 keyframes based on the video context; (2) \textit{w/ 8 frames}: Select 8 keyframes with the same input format as most open-source MLLMs; (3) \textit{w/ captions}: Replace 32 keyframes with its corresponding captions generated by GPT-4o; (4) w/ only-qa: Input only the question without any frames or captions.

\paragraph{Open-Source Image-based MLLMs.}
We select several image-based MLLMs that demonstrated strong performance in EgoThink~\citep{Cheng_2024_CVPR}.
Below, we provide a brief introduction and present their detailed component information in Table~\ref{tab:sta_mllms}.

\begin{itemize}[leftmargin=*]
    % \item \textbf{LLaVA}~\citep{zhang2023llava} uses linear layers to connect the vision encoder and LLM, training the projection matrix and pre-trained language model to enhance adaptability.
    \item \textbf{mPLUG-Owl2}~\citep{ye2024mplug} integrates shared functional modules to promote modality collaboration and includes a modality-adaptive module to preserve modality-specific features.
    \item \textbf{Qwen-VL-Chat}~\citep{bai2023qwen} employs a single-layer cross-attention with random initialization, trained with approximately 1.5 billion image-text pairs, and aligns with human interaction.
    \item \textbf{LLaVA-1.5}~\citep{liu2024improved} utilizes academic task data and replaces the linear visual language connector with a two-layer MLP connector.
    \item \textbf{LLaMA-Adapter V2}~\citep{gao2023llama} proposes an early fusion strategy that effectively adapts LLaMA~\citep{touvron2023llama} to visual instruction models.
    % \item \textbf{Idefics2}~\citep{laurenccon2024matters} is a lightweight model adaptable to different image resolutions, significantly improved in document understanding, OCR, and visual reasoning.
    % \item InternVL-Chat-v1.5~\citep{chen2024far} uses a strong visual encoder and dynamic high-resolution method to enhance visual performance, and uses high-quality bilingual dataset to enhance Chinese and OCR-related capabilities.
\end{itemize}

\paragraph{Open-Source Video-based MLLMs.} 
We comprehensively select the most popular and high-performance video-based MLLMs to evaluate. Our video-based MLLMs include a series of general models and three grounding-specific models. The detailed components are shown in Table~\ref{tab:sta_mllms}.
\begin{itemize}[leftmargin=*, itemsep=1pt, topsep=0pt]
    \item \textbf{LWM}~\citep{liu2024worldmodelmillionlengthvideo} uses Blockwise RingAttention and masked sequence packing to manage long video and language sequences, enabling training on contexts up to 1 million tokens for better multimodal understanding.
    \item \textbf{TimeChat}~\citep{ren2024timechat} is a time-sensitive multimodal large language model that aligns visual information with specific time frames. It utilizes a sliding video Q-Former to adapt to videos of varying lengths.
    \item \textbf{GroundingGPT}~\citep{li2024groundinggpt} effectively enhances the understanding and grounding of fine-grained image, video, and audio modalities through a three-stage, coarse-to-fine training strategy.
    \item \textbf{InternVL2}~\citep{chen2023internvl} builds on InternVL's QLLaMA progressive alignment strategy. It optimizes vision-language alignment while scaling up the language model in stages, starting small and expanding gradually, with data refined from coarse to fine.  
    \item \textbf{InternLM-XComposer-2.5}~\citep{zhang2024internlm} introduces RoPE extrapolation for long-context handling, ultra-high resolution understanding, fine-grained video comprehension, and multi-turn multi-image dialogue, and extra LoRA parameters for advanced text-image composition.
    \item \textbf{Video-LLaVA}~\citep{lin2023video} proposes a unified visual representation method that aligns images and videos within the language feature space. This approach enhances multimodal interactions and leverages a mixed dataset of images and videos to mutually improve each modality.
    \item \textbf{PG-Video-LLaVA}~\citep{munasinghe2023pg} is a video-based MLLM with pixel-level grounding capabilities. It can also integrate audio to enhance video understanding. Additionally, its modular design enhances flexibility.
    \item \textbf{mPLUG-Owl3}~\citep{ye2024mplugowl3longimagesequenceunderstanding} introduces hyper attention blocks to efficiently integrate vision and language into a shared semantic space, improving long image sequence processing.    
    \item \textbf{MiniCPM-V-2.6}~\citep{yao2024minicpmvgpt4vlevelmllm} utilizes the adaptive visual encoding mechanism of LLaVA-UHD~\citep{xu2024llava} and various end-side optimizations to compress the multimodal model. 
    \item \textbf{Qwen2-VL}~\citep{Qwen-VL, Qwen2-VL} has been upgraded with Naive Dynamic Resolution and Multimodal Rotary Position Embedding (M-ROPE) technologies, improving its multimodal data processing and understanding capabilities.
\end{itemize}

\begin{table}[t!]
    \centering
    \caption{Experimental results of video question answering. OE, OO, OI, OC, OS, OP denote object existence, object order, object interaction, object count, object state, object prediction. AE, AS, AC indicates action existence, action sequence, action count. SE, ST, SP denote scene existence, scene transition, scene prediction. The \textbf{bold} font denotes the best performance and the \underline{underline} font denotes the second-best performance.}
    \label{tab:result_vqa}
    \resizebox{\textwidth}{!}{
    \begin{tabular}{lccccccccccccc}
    \toprule
        \multirow{2}{*}{\textbf{Models}} & \multicolumn{6}{c}{\textbf{Object}} & \multicolumn{3}{c}{\textbf{Action}} & \multicolumn{3}{c}{\textbf{Scene}} & \multirow{2}{*}{\textbf{Average}} \\
        \cmidrule(rl){2-7}
        \cmidrule(rl){8-10}
        \cmidrule(rl){11-13}
        & OE & OO & OI & OC & OS & OP & AE & AS & AC & SE & ST & SP \\
        \midrule
        % \textbf{GPT-4-turbo} {\footnotesize w/ narr.} \\
        \textbf{GPT-4o} {\footnotesize w/ only-qa} & 13.00 & \textcolor{white}{0}0.00 & 12.00 & \textcolor{white}{0}6.00 & 31.00 & 23.00 & 25.00 & \textcolor{white}{0}4.00 & \textcolor{white}{0}2.00 & 18.00 & \textcolor{white}{0}6.00 & \underline{20.00}  & 13.33 \\
        \textbf{GPT-4o} {\footnotesize w/ captions} & \underline{51.00} & 16.00 & 14.00 & 30.00 & 25.00 & \underline{44.00} & 34.00 & \textcolor{white}{0}5.00 & 22.00 & 42.00 & \textbf{28.00} & 16.00 & 27.25 \\
        \textbf{GPT-4o} {\footnotesize w/ 8 frames} & \underline{51.00} & 16.00 & \textbf{30.00} & 33.00 & \textbf{35.00} & \textbf{45.00} & \underline{38.00} & \textbf{25.00} & 22.00 & 43.00 & 23.00 & \textbf{24.00} & \textbf{32.83}\\
        \textbf{GPT-4o} {\footnotesize w/ 32 frames} & \textbf{52.00} & \underline{18.00} & \textbf{30.00} & \textbf{35.00} & 32.00 & 40.00 & \textbf{39.00} & \underline{20.00} & 24.00 & \underline{46.00} & 20.00 & 18.00 & \underline{31.17} \\
        \midrule
        \textbf{mPLUG-Owl2-llama2-7B} & 29.00 &  \textcolor{white}{0}6.00 & 15.00 & 30.00 & 10.00 & 16.00 & 28.00 & \textcolor{white}{0}8.00 & 28.00 & 20.00 & 10.00 & \textcolor{white}{0}6.00 & 17.17 \\
        \textbf{Qwen-VL-7B-Chat} & 41.00 & \textcolor{white}{0}7.00 & 13.00 & 33.00 & 14.00 & 30.00 & 17.00 & \textcolor{white}{0}3.00 & 27.00 & 16.00 & 13.00 & 10.00 & 18.67 \\
        \textbf{LLaVA-1.5-7B} & 46.00 & \textcolor{white}{0}7.00 & 17.00 & \underline{34.00} & 22.00 & 24.00 & 25.00 & \textcolor{white}{0}1.00 & 14.00 & 20.00 & 13.00 & 16.00 & 19.92 \\
        \textbf{LLaMA-Adapter-V2-7B} & 48.00 &  \textcolor{white}{0}5.00 & 26.00 & 17.00 & 19.00 & 39.00 & 14.00 & \textcolor{white}{0}9.00 & 35.00 & 24.00 & 10.00 & 16.00 & 21.80\\
        % Idefics2 \\
        \midrule
        \textbf{LWM-Chat-32k-Jax-7B} & 42.00 & \textcolor{white}{0}3.00 & 20.00 & 12.00 & 10.00 & 11.00 & 20.00 & \textcolor{white}{0}4.00 & 21.00 & 27.00 & \textcolor{white}{0}9.00 & \textcolor{white}{0}5.00 & 15.33\\
        \textbf{TimeChat-7B} & 42.00 & \textcolor{white}{0}5.00 & 15.00 & 21.00 & 11.00 & 23.00 & 20.00 & \textcolor{white}{0}4.00 & 20.00 & 31.00 & 14.00 & 14.00 & 18.33 \\
        \textbf{GroundingGPT-7B} & 43.00 & \textcolor{white}{0}3.00 & 20.00 & 30.00 & 10.00 & 23.00 & 22.00 & \textcolor{white}{0}4.00 & 32.00 & 23.00 & 19.00 & 14.00 & 20.25 \\
        \textbf{InternVL2-8B} & 43.00 & 16.00 & 21.00 & 18.00 & 20.00 & 27.00 & 19.00 & \textcolor{white}{0}4.00 & 15.00 & 37.00 & 17.00 & 12.00 & 20.75 \\
        \textbf{InternLM-XComposer2.5-7B} & 36.00 & \textcolor{white}{0}6.00 & 24.00 & 22.00 & 19.00 & 34.00 & 30.00 & \textcolor{white}{0}2.00 & 30.00 & 31.00 & 11.00 & 12.00 & 21.42 \\
        \textbf{Video-LLaVA-7B} & 44.00 & \textcolor{white}{0}8.00 & 19.00 & \underline{34.00} & 15.00 & 30.00 & 18.00 & \textcolor{white}{0}3.00 & \textbf{38.00} & 28.00 & 11.00 & 11.00 & 21.58 \\
        \textbf{PG-Video-LLaVA-7B} & 49.00 & \textcolor{white}{0}5.00 & 21.00 & 15.00 & 23.00 & 37.00 & 25.00 & \textcolor{white}{0}3.00 & 16.00 & 35.00 & 18.00 & 20.00 & 22.25 \\
        \textbf{mPLUG-Owl3-7B} & 32.00 & \textcolor{white}{0}7.00 & 26.00 & 13.00 & \underline{33.00} & 34.00 & 18.00 & \textcolor{white}{0}6.00 & \underline{36.00} & 37.00 & 23.00 & 10.00 & 22.92\\
        \textbf{MiniCPM-V-2.6-8B} & 48.00 & 12.00 & \underline{28.00} & 16.00 & 25.00 & 42.00 & 31.00 & 11.00 & 15.00 & 42.00 & 23.00 & 18.00 & 25.92\\
        \textbf{Qwen2-VL-7B-Instruct} & 36.00 & \textbf{19.00} & \underline{28.00} & 28.00 & 28.00 & 43.00 & 24.00 & \textcolor{white}{0}9.00 & 20.00 & \textbf{48.00} & \underline{24.00} & \underline{20.00} & 27.25\\
    \bottomrule
    \end{tabular}}
\end{table}

\subsection{Results}

\noindent\textbf{Video Question-Answering.}
The results of the video question-answering task are shown in Table~\ref{tab:result_vqa} and Table~\ref{tab:result_others}.
MLLMs perform poorly, with a best average accuracy of 32.82\% across all dimensions (35.00\% for object, 28.33\% for action, and 26.33\% for scene elements), indicating struggles with egocentric video question-answering.
GPT-4o with 8 frames performs better than with 32 frames but still underperforms compared to some open-source video MLLMs in certain dimensions.
Two probable reasons are: (1) GPT-4o's sensitivity to privacy policies for indoor videos, causing it to refuse more questions given more images; (2) insufficient information from extracted keyframes.
GPT-4o with captions sometimes matches or surpasses the 8 or 32-frame setups in scene transitions, but performs poorly in object interaction and action sequence dimensions, indicating that captions provide sufficient high-level abstraction but lack detailed low-level action information.
We regard the GPT-4o with only-qa as a baseline to demonstrate state-of-the-art performance using only question-answering pairs without any vision information. All other MLLMs perform better than the average accuracy of GPT-4o with only-qa, showing that our benchmark indeed requires vision information to solve these problems.
Open-source video-based MLLMs generally surpass image-based MLLMs, highlighting the need for full video information, especially in dynamic dimensions.
Among these, Qwen2-VL-7B-Instruct achieves the best performance, even surpassing GPT-4o in two dimensions and achieving the second-best performance in three dimensions.

\noindent\textbf{Hierarchy Planning.}
The hierarchy planning results are shown in Table~\ref{tab:result_others}, with the average video duration being 1008.26 seconds.
In the High-to-Mid task, GPT-4o series models and image-based MLLMs, which process multiple or single images, lack sufficient information to determine the entire progress and predict the next step. 
Hence, increasing the total number of frames significantly improves performance.
For video-based models, the best performance of MiniCPM is comparable to the state-of-the-art performance of GPT-4o with 32 frames but still performs poorly, indicating significant room for improvement.
% In this version, we only assess the performance to ask API-based models to determine the score based on given references. However, the next step to achieve a goal is open-vocabulary in the real world, we will explore a better evaluation method in the future.
For the Mid-to-Low task, the most notable phenomenon is that GPT-4o series models significantly outperform open-source MLLMs, which achieve about 0.05 accuracy. 
The main reason behind this phenomenon is the limited long-context capability and instruction-following capability of open-source MLLMs.
We can only provide them with a compressed function document, and they often do not generate answers following the output format.

\noindent\textbf{Visual Grounding.}
Visual grounding tasks involve identifying specific objects, frames, or temporal segments within a video. API-based and image-based MLLMs abandon this information after extracting keyframes, necessitating the use of open-source video-based MLLMs for performance assessment.
Due to the new design of object and frame grounding tasks, these MLLMs are not yet optimized for these formats, leading to generally poor performance.
It is understandable that object grounding in a single image remains a challenging task for MLLMs, even more so within a video context.
For temporal grounding, some MLLMs especially trained for this task achieve relatively high scores, with PG-Video-LLaVA scoring 16.18.
Surprisingly, MiniCPM performs well across all grounding dimensions, despite not being specially trained for these tasks.
Although the performances of MLLMs are poor, we believe these tasks have a significant impact on downstream tasks in Embodied AI and deserve more attention.

\noindent\textbf{Reward Modeling.}
As shown in Table~\ref{tab:result_others}, the critique task is a binary classification task with a random guess baseline of 50\%. Therefore, the overall performance of MLLMs is suboptimal, with the best accuracy reaching only 59.39\%, indicating that MLLMs struggle to determine whether a task has been completed.
For the feedback task, GPT-4o with 8 frames (33.46\%) and 32 frames (34.64\%) significantly outperforms the best results from other API-based methods (14.58\%) and open-source MLLMs (13.09\%).
This demonstrates that generating feedback requires more fine-grained visual information not present in captions and superior reasoning capability.

\begin{table}[t!]
    \centering
    \vspace{-5mm}
    \caption{Experimental results of video question answerng, hierarchy planning, visual grounding, and reward modeling tasks. The \textbf{bold} font denotes the best performance and the \underline{underline} font denotes the second-best performance.}
    \resizebox{\textwidth}{!}{
    \begin{tabular}{lcccccccccc}
    \toprule
        \multirow{2}{*}{\textbf{Models}} & \multicolumn{3}{c}{\textbf{Video Question Answering}} & \multicolumn{2}{c}{\textbf{Hierarchy Planning}} & \multicolumn{3}{c}{\textbf{Visual Grounding}} & \multicolumn{2}{c}{\textbf{Reward Modeling}}  \\
        \cmidrule(rl){2-4}
        \cmidrule(rl){5-6}
        \cmidrule(rl){7-9}
        \cmidrule(rl){10-11}
        & Object & Action & Scene & High-to-Mid & Mid-to-Low & Object & Frame & Temporal & Critique & Feedback \\ 
        \midrule
        \textbf{GPT-4o} {\footnotesize w/ only-qa} & 14.17 & 10.33 & 14.67 & \textcolor{white}{0}8.86 & 32.56 & - & - & - & 48.46 & \textcolor{white}{0}6.81 \\
        \textbf{GPT-4o} {\footnotesize w/ captions} & 30.00 & 20.33 & \underline{28.67} & \textcolor{white}{0}9.53 & 33.65 & - & - & - & \underline{58.82} & 14.58 \\
        \textbf{GPT-4o} {\footnotesize w/ 8 frames} & \textbf{35.00} & \textbf{28.33} & \textbf{30.00} & 12.04 & \textbf{35.47} & - & - & - & 58.74 & \underline{33.46}  \\
        \textbf{GPT-4o} {\footnotesize w/ 32 frames} & \underline{34.50} & \underline{27.67} & 26.33 & \textbf{14.97} & \underline{35.08} & - & - & - & \textbf{59.39} & \textbf{34.64} \\  
        \midrule
        \textbf{mPLUG-Owl2-llama2-7B} & 17.67 & 21.33 & 12.00 &  \textcolor{white}{0}5.77 &  \textcolor{white}{0}0.00  & - & - & - & 41.26 & \textcolor{white}{0}1.56\\
        \textbf{Qwen-VL-7B-Chat} & 23.00 & 15.67 & 13.00 & 10.79 & \textcolor{white}{0}0.04 & - & - & - & 49.19 & \textcolor{white}{0}4.08 \\
        \textbf{LLaVA-1.5-7B} & 25.00 & 13.33 & 16.33 & \textcolor{white}{0}2.59 &\textcolor{white}{0}0.01& - & - & - & 53.72 & \textcolor{white}{0}3.53 \\
        \textbf{LLaMA-Adapter-V2-7B} & 25.67 & 19.33 & 16.67 & \textcolor{white}{0}4.59 & \textcolor{white}{0}0.03 & - & - & - & 39.64 & \textcolor{white}{0}2.89 \\
        % Idefics2 & & & & & & & & & & \\
        \midrule
        \textbf{LWM-Chat-32k-Jax-7B} & 16.33 & 15.00 & 13.67 & \textcolor{white}{0}1.33 & \textcolor{white}{0}0.00 & \textcolor{white}{0}0.00 & \textcolor{white}{0}0.00 & \textcolor{white}{0}0.00 & 22.09 & \textcolor{white}{0}0.00   \\
        \textbf{TimeChat-7B} & 19.50 & 14.67 & 19.67 & \textcolor{white}{0}3.85 & \textcolor{white}{0}0.01 & \textcolor{white}{0}0.00 & \textcolor{white}{0}0.00 & \underline{14.56} & 47.25 & \textcolor{white}{0}0.57\\
        \textbf{GroundingGPT-7B} & 21.50 & 19.33 & 18.66 & \textcolor{white}{0}5.69 & \textcolor{white}{0}0.05  &  \textcolor{white}{0}\textbf{0.76} & \textcolor{white}{0}\underline{0.54} & \textcolor{white}{0}0.44 & 51.13 & \textcolor{white}{0}2.19 \\
        \textbf{InternVL2-8B}  & 24.17 & 12.67 & 22.00 & \textcolor{white}{0}3.34 & \textcolor{white}{0}0.05 & \textcolor{white}{0}0.09  & \textcolor{white}{0}0.00 & \textcolor{white}{0}6.87 & 52.67 & \textcolor{white}{0}0.71 \\
        \textbf{InternLM-XComposer2.5-7B} & 23.50 & 20.67 & 18.00 & \textcolor{white}{0}9.62 & \textcolor{white}{0}0.04 & \textcolor{white}{0}0.00 & \textcolor{white}{0}\underline{0.54} & \textcolor{white}{0}3.50 & 51.41 & \textcolor{white}{0}8.23 \\
        % Video-LLaVA-7B & 25.00 & 19.67 & 16.67 & & & & & & 47.94 &  \\
        \textbf{PG-Video-LLaVA-7B} &25.00 & 14.67 & 24.33 &  \textcolor{white}{0}5.35 &\textcolor{white}{0}0.00 & \textcolor{white}{0}0.08 & \textcolor{white}{0}0.00 & \textbf{16.18} & 48.30 & \textcolor{white}{0}6.27 \\
        \textbf{mPLUG-Owl3-7B} & 24.17 & 20.00 & 23.33 & 12.29 & \textcolor{white}{0}0.03 & \textcolor{white}{0}0.00 & \textcolor{white}{0}0.00 & \textcolor{white}{0}0.00 & 50.00 & \textcolor{white}{0}9.09 \\
        \textbf{MiniCPM-V-2.6-8B} & 28.50 & 19.00 & 27.67 & \underline{14.13} & \textcolor{white}{0}0.06 & \textcolor{white}{0}\underline{0.35} & \textcolor{white}{0}\textbf{1.63 }& 11.30 & 51.54 & 13.09\\
        \textbf{Qwen2-VL-7B-Instruct} & 30.33 & 16.00 & 27.67 & \textcolor{white}{0}9.88& \textcolor{white}{0}0.00 & \textcolor{white}{0}0.00 & \textcolor{white}{0}0.00 & \textcolor{white}{0}0.00 & 49.03 & \textcolor{white}{0}4.62 \\
    \bottomrule
    \end{tabular}}
    \label{tab:result_others}
\end{table}

\section{Conclusion}
\label{sec:limitation}

% MLLMs have the potential to become a new paradigm for Embodied AI, particularly in high-level perception and cognition.
In this paper, we introduce VidEgoThink, a comprehensive benchmark designed to evaluate egocentric video understanding across four critical functions in Embodied AI.
Our assessment of popular API-based and open-source MLLMs reveals that these models still face significant challenges in processing egocentric videos.
Although GPT series models perform relatively better, they exhibit notable deficiencies in certain areas, highlighting the need for further improvements and optimizations.
VidEgoThink underscores the limitations of current MLLMs in handling first-person perspective data, thereby indicating directions for future research and advancements
% , especially with the rapid development of humanoid  robotics and wearable glasses.

\noindent\textbf{Limitations.}
VidEgoThink is the first to propose four tasks for assessing egocentric video understanding in MLLMs for Embodied AI. 
However, it has limited data diversity and immature evaluation methods, particularly in hierarchy planning and reward modeling. 
Future work should improve these aspects and address the high costs of human annotation and API-based evaluations, which limit the number of question-answer pairs.
We plan to expand the benchmark and develop egocentric foundation models for robotics.

\noindent\textbf{Broader Impacts.}
Two key areas for the future of Embodied AI are \textit{egocentric video} and \textit{multi-modal large language models}.
On the one hand, our real world cannot be mapped to virtual simulators exactly the same way.
Real-world environments cannot be exactly replicated in virtual simulators, making egocentric video a preferred method for collecting action data, especially with the rise of smart glasses and humanoid robots.
Learning from egocentric video is crucial for future advancements.
Although end-to-end MLLMs for Embodied AI are still an open research question, we believe a hierarchical system that uses vision-language models for perception and cognition is an emerging paradigm. Ideal foundation models should function in the real world, capable of thinking, understanding, and interacting like humans.


% Yuyang You. Xiang Yue, Yuanzhi Li, and Jiangjie Chen for their early discussion.

\bibliography{iclr2024_conference}
\bibliographystyle{iclr2024_conference}

\clearpage
\appendix
\section{Prompt Hubs}
\label{app:prompt}

To address concerns about potential data breaches through prompts, here we only release the detailed prompts for each task to facilitate inference and evaluation.

\subsection{Model Inference Prompts}

As an example, we list the general prompts for 8 frames, 32 frames and open-source MLLMs.
The inference type of ``caption'' for GPT series models will add a prompt ``\textit{Here is the captions of the video: \{caption\}.}'' after the sentence ``\textit{Imagine you are the camera wearer (I) who recorded the video}''.
For the inference type of ``only-qa'', we delete the prompt ``\textit{Imagine you are the camera wearer (I) who recorded the video}''.

\begin{itemize}[leftmargin=*, itemsep=1pt, topsep=0pt]
    \item \textbf{Video Question Answering:} \textit{Imagine you are the camera wearer (I) who recorded the video. Please directly answer the question as short as possible. Question: \{question\} Short answer:}
    \item \textbf{High-to-Mid in Hierarchy Planning:} \textit{Imagine you are the camera wearer (I) who recorded the video. Given the high-level goal (e.g., 'making dumpling') and the current progress video, you need to predict the next mid-level step (e.g., fold dumplings on a cutting board) to achieve the goal. Please directly generate the next one step as short as possible. Question: \{question\} Short answer:}
        \item \textbf{Mid-to-Low in Hierarchy Planning:} \textit{Imagine you are the camera wearer (I) who recorded the video. Here are a set of actionable functions below.\\
    \textnormal{[begin of actionable function and documentation]}\\
    \{`put': `put(\textless arg1\textgreater, \textless arg2\textgreater) is used to place an object at a specified or default location. \textless arg1\textgreater refers to the item to be placed, whereas \textless arg2\textgreater is optional and specifies the location where the item should be placed. If \textless arg2\textgreater is omitted, the item is placed in a generic, predefined area.',\\
    `grab': `grab(\textless arg1\textgreater, \textless arg2\textgreater) is used to simulate the action of grasping or picking up objects, especially in a kitchen setting. \textless arg1\textgreater refers to the primary object to be grabbed, while \textless arg2\textgreater is optional and denotes an associated tool or container that aids in handling or processing the primary object.', \\ 
    `talk': `talk(\textless arg1\textgreater, \textless arg2\textgreater) is used to simulate a conversation scenario with specific entities. \textless arg1\textgreater is mandatory and specifies the primary entity involved in the conversation, such as a 'woman', 'man', or 'person'. \textless arg2\textgreater is optional and typically represents a secondary entity or context within the conversation, providing additional detail or focus.',\\
    `close': `close(\textless arg1\textgreater, \textless arg2\textgreater) is used to encapsulate or seal an item, either partially or completely. \textless arg1\textgreater refers to the object to be closed or covered, and \textless arg2\textgreater is optional, describing the material or object used for closing or covering \textless arg1\textgreater. If \textless arg2\textgreater is omitted, the closing is done without any specified covering.',\\
    `adjust': `adjust(\textless arg1\textgreater, \textless arg2\textgreater) is used to modify the position or settings of objects or items. \textless arg1\textgreater is mandatory and specifies the primary object to adjust, while \textless arg2\textgreater is optional and used for adjustments involving a specific secondary object or location relative to the first.',\\
    `arrange': `arrange(\textless arg1\textgreater, \textless arg2\textgreater) is used to organize objects systematically within a predefined space. \textless arg1\textgreater refers to the items to be arranged, while \textless arg2\textgreater is optional and specifies the area or container where these items will be organized. If \textless arg2\textgreater is omitted, the items are arranged in a default designated space.',\\
    `open': `open(\textless arg1\textgreater, \textless arg2\textgreater) is used to manipulate the state of various containers or coverings by opening them. \textless arg1\textgreater refers to the primary object or container that needs to be opened, like a 'pot' or 'drawer'. \textless arg2\textgreater is optional and specifies a secondary descriptor or specific part of the primary object, like 'top' or 'front', indicating a particular method or area of opening.',\\
    `walk': `walk(\textless arg1\textgreater, \textless arg2\textgreater) is used to move an entity towards a specified location within an environment. \textless arg1\textgreater refers to the primary location or object the entity should head towards, and \textless arg2\textgreater refers to optional additional parameters that provide extra directional or contextual details to refine the movement.',\\
    `empty': 'empty(\textless arg1\textgreater, \textless arg2\textgreater) is used to transfer a specified item from one holding medium to another specified container. \textless arg1\textgreater refers to the item being transferred, while \textless arg2\textgreater is the destination container where the item is moved to.',\\
    `move': 'move(\textless arg1\textgreater, \textless arg2\textgreater) is used to transfer items from one place to another. \textless arg1\textgreater refers to the item that is being moved. \textless arg2\textgreater is optional and specifies where the item should be placed; if omitted, it indicates the item is moved without a specific destination in mind, likely for clearing space or as an intermediate step.',\\
    `push': 'push(\textless arg1\textgreater, \textless arg2\textgreater) is used to initiate a push action on various objects or elements. \textless arg1\textgreater refers to the main object or element to be pushed, and \textless arg2\textgreater is optional and used to specify a particular part or aspect of \textless arg1\textgreater for a more precise push action.',\\
    `clean': 'clean(\textless arg1\textgreater, \textless arg2\textgreater) is used to cleanse various items, which may include food or non-food objects. \textless arg1\textgreater refers to the primary item that requires cleaning, while \textless arg2\textgreater is optional and specifies additional items or the context like the cleaning environment or method. If \textless arg2\textgreater is omitted, the function adapts its operation to effectively clean \textless arg1\textgreater alone.',\\
    `rotate': `rotate(\textless arg1\textgreater, \textless arg2\textgreater) is used to turn or move an item, typically in a culinary context. \textless arg1\textgreater refers to the item that needs to be rotated. \textless arg2\textgreater is optional and describes the utensil or tool used to facilitate the rotation. If \textless arg2\textgreater is omitted, the item is rotated manually or with a default method.',\\
    `serve': ``serve(\textless arg1\textgreater, \textless arg2\textgreater) is used to manage the distribution or placement of items. \textless arg1\textgreater refers to the item to be served or used, and \textless arg2\textgreater is optional, indicating the person or the hand (right or left) that will handle the item. If \textless arg2\textgreater is omitted, the item is handled by default means.',\\
    `shell': 'shell(\textless arg1\textgreater, \textless arg2\textgreater) is used to remove the outer covering from items, typically food-related like seeds, vegetables, and fruits. \textless arg1\textgreater is mandatory and specifies the item from which the shell or outer layer needs removal. \textless arg2\textgreater is optional and indicates any tool that might assist in the shelling process, such as a knife or fork. If \textless arg2\textgreater is omitted, the item is shelled using standard methods.',\\
    `turn\_on': `turn\_on(\textless arg1\textgreater, \textless arg2\textgreater, etc) is used to activate one or multiple household or industrial appliances. \textless arg1\textgreater is mandatory and refers to the primary appliance that needs to be activated. \textless arg2\textgreater, etc, represent additional appliances that can be optionally activated simultaneously.',\\
    `turn\_off': `turn\_off(\textless arg1\textgreater) is used to deactivate various devices or utilities. \textless arg1\textgreater refers to the object or device to be deactivated, such as a 'socket', 'tap', or 'blending machine'.',\\
    `cut': `cut(\textless arg1\textgreater, \textless arg2\textgreater) is used to perform the action of cutting on various items. \textless arg1\textgreater refers to the item to be cut, which is mandatory. \textless arg2\textgreater is optional and denotes the tool used for cutting; if \textless arg2\textgreater is omitted, a standard cutting tool is assumed.',\\
    `throw': `throw(\textless arg1\textgreater, \textless arg2\textgreater) is used to dispose of or place an object in a specified or default location. \textless arg1\textgreater refers to the item to be disposed of or relocated, whereas \textless arg2\textgreater is optional and designates the location where the item should be placed. If \textless arg2\textgreater is omitted, the function selects a default disposal method or location based on the item or context.',\\
    `mix': `mix(\textless arg1\textgreater, \textless arg2\textgreater) is used to combine or stir ingredients, typically in a cooking context. \textless arg1\textgreater refers to the item or ingredients to be mixed, and \textless arg2\textgreater is optional and denotes the tool used for mixing, such as a spoon or paddle. When \textless arg2\textgreater is omitted, the method of mixing is unspecified or assumed to be manual.',\\
    `touch': `touch(\textless arg1\textgreater, \textless arg2\textgreater) is used to simulate the action of touching various items or materials. \textless arg1\textgreater refers to the object or material that is the primary focus of the action, whereas \textless arg2\textgreater is optional and provides additional detail on a specific part of the item to touch, assuming a generic aspect if omitted.',\\
    `eat': `eat(\textless arg1\textgreater, \textless arg2\textgreater) is used to perform the action of consuming a specified item. \textless arg1\textgreater refers to the item to be consumed. \textless arg2\textgreater is optional and describes the method by which the food is to be eaten, for example, 'slowly'.',\\
    `pull': `pull(\textless arg1\textgreater, \textless arg2\textgreater) is used to simulate the action of pulling something within a specific context. \textless arg1\textgreater refers to the object that is being pulled, such as a drawer or an oven grill. \textless arg2\textgreater is optional and describes a secondary reference or location, like a pan or a steel cabinet, which adds context to where the object is located or what it is associated with. If \textless arg2\textgreater is omitted, the action focuses solely on \textless arg1\textgreater.',\\
    `unfold': `unfold(\textless arg1\textgreater, \textless arg2\textgreater=None) is used to expand or open various types of items. \textless arg1\textgreater refers to the item to be unfolded, such as fabric, body parts, or food items. \textless arg2\textgreater is optional and allows for additional specifications on how the unfolding should be performed, tailored based on the nature of the item. If \textless arg2\textgreater is omitted, basic operations are performed.',\\
    `dip': `dip(\textless arg1\textgreater, \textless arg2\textgreater) is used to immerse an item into a container. \textless arg1\textgreater refers to the item to be dipped, such as 'dough' or 'hand', and \textless arg2\textgreater describes the container like 'bowl of water' or 'flour'. This function facilitates operations involving coating or soaking an item.',\\
    `observe': `observe(\textless arg1\textgreater) is used to examine the specified environment or objects. \textless arg1\textgreater refers to an array containing one or more strings that describe what should be focused on during the observation. At least one string is mandatory to define the scope of observation, while additional strings are optional to provide more detail.',\\
    `taste': `taste(\textless arg1\textgreater, \textless arg2\textgreater) is used to simulate the action of tasting a specified item with or without a utensil. \textless arg1\textgreater refers to what is being tasted, such as food or soup. \textless arg2\textgreater is optional and specifies the utensil used for tasting, like a spoon. If \textless arg2\textgreater is omitted, the action of tasting is assumed to be done without any specific utensil.',\\
    `apply': `apply(\textless arg1\textgreater, \textless arg2\textgreater) is used to perform operations involving the application or manipulation of cooking ingredients or tools. \textless arg1\textgreater refers to the primary material or tool being used, such as 'flour' or 'oil'. \textless arg2\textgreater is optional and typically refers to the target where \textless arg1\textgreater is applied, like 'dough' or 'frying pan'.',\\
    `switch': `switch(\textless arg1\textgreater) is used to change or replace the current tool in use within a system or application. \textless arg1\textgreater corresponds to the name of the tool that the function will switch to.',\\
    `roll': `roll(\textless arg1\textgreater, \textless arg2\textgreater) is used to flatten or shape an item using a tool. \textless arg1\textgreater refers to the item to be rolled, such as dough or foil. \textless arg2\textgreater is optional and indicates the tool used for rolling, like a 'rolling pin' or 'rolling board'. If \textless arg2\textgreater is not specified, a default tool or method is used to roll \textless arg1\textgreater.',\\
    `lay': `lay(\textless arg1\textgreater, \textless arg2\textgreater) is used to place objects or substances within a specific environment or a default setting if not specified. \textless arg1\textgreater refers to what is being placed, and \textless arg2\textgreater is optional and defines where the item is placed.',\\
    `gesture': `gesture(\textless arg1\textgreater, \textless arg2\textgreater, etc) is used to perform low-level actions based on the type of gesture or action specified. \textless arg1\textgreater is mandatory and refers to the string specifying the type of gesture or action to be executed. \textless arg2\textgreater is optional and allows for additional details or modifications to the gesture when necessary.',\\
    `steer': `steer(\textless arg1\textgreater, \textless arg2\textgreater, etc) is used to manipulate or interact with an object in a controlled environment. \textless arg1\textgreater refers to any object that requires handling or operation. \textless arg2\textgreater is optional, enhancing or specifying the nature of the interaction.',\\
    `operate': `operate(\textless arg1\textgreater, \textless arg2\textgreater) is used to activate or manage a specified device. \textless arg1\textgreater refers to the name of the device being operated, while \textless arg2\textgreater is optional and allows specific operational parameters to be passed, such as temperature, duration, or intensity.',\\
    `store': `store(\textless arg1\textgreater, \textless arg2\textgreater) is used to log or record items into a storage system. \textless arg1\textgreater refers to the list of items to be stored, which can include a single item or multiple items listed together. \textless arg2\textgreater is optional and specifies where the items are to be stored, indicating the physical or logical grouping.',\\
    `tilt': `tilt(\textless arg1\textgreater, \textless arg2\textgreater) is used to tip or angle an item, often to enable actions like pouring. \textless arg1\textgreater refers to the item that needs to be tilted. \textless arg2\textgreater is optional and defines the degree or direction of tilt. If \textless arg2\textgreater is omitted, a default tilt setting is used.',\\
    `lift': `lift(\textless arg1\textgreater, \textless arg2\textgreater) is used to simulate the action of picking up or lifting an object or a group of objects. \textless arg1\textgreater refers to the primary object to be lifted, and \textless arg2\textgreater is optional, indicating an additional item or tool used alongside the primary object during the lifting process.',\\
    `scrape': `scrape(\textless arg1\textgreater, \textless arg2\textgreater) is used to perform the action of scraping one item against another. \textless arg1\textgreater refers to the item to be scraped, which is mandatory, such as 'cabbage' or 'vegetables'. \textless arg2\textgreater is optional and refers to the surface or tool against which the item is scraped, like 'board' or 'frying pan'. If \textless arg2\textgreater is omitted, the function defaults to a generic, predefined scraping context.',\\
    `bend': `bend(\textless arg1\textgreater, \textless arg2\textgreater, \textless arg3\textgreater) is used to modify the shape or structure of an object. \textless arg1\textgreater refers to the object undergoing the bending. \textless arg2\textgreater and \textless arg3\textgreater are optional and specify the degree and the direction of the bend, respectively, allowing for precise control over the bending process.',\\
    `hit': `hit(\textless arg1\textgreater, \textless arg2\textgreater) is used to simulate the action of one object striking another. \textless arg1\textgreater refers to the primary object being hit, while \textless arg2\textgreater is optional and indicates any additional object used in the hitting action, such as a tool.',\\
    `reduce\_heat': `reduce\_heat(\textless arg1\textgreater, \textless arg2\textgreater) is used to lower the temperature or heat output of a specific device. \textless arg1\textgreater refers to the device on which the heat reduction is to be applied, and \textless arg2\textgreater is optional and provides an interface or method for achieving the heat reduction, allowing for precise control when specified.',\\
    `rub': `rub(\textless arg1\textgreater, \textless arg2\textgreater) is used to simulate the action of rubbing an object or surface. \textless arg1\textgreater refers to the primary object that is being rubbed, and \textless arg2\textgreater is optional, referring to a secondary object or surface involved in the rubbing, which can enhance or alter the rubbing context. If \textless arg2\textgreater is omitted, the rubbing action is considered to be performed solely with \textless arg1\textgreater.',\\
    `add': `add(\textless arg1\textgreater, \textless arg2\textgreater) is used to simulate placing an item into a container or context within a simulated environment. \textless arg1\textgreater refers to the object to be added, which is mandatory. \textless arg2\textgreater is optional and specifies the location or receptacle for the item. If \textless arg2\textgreater is omitted, the item is added to a default location or context.',\\
    `mould': `mould(\textless arg1\textgreater) is used to shape or form a material into a desired structure. \textless arg1\textgreater refers to the substance that needs to be shaped, such as clay, dough, or plastic.',\\
    `knead': `knead(\textless arg1\textgreater, \textless arg2\textgreater) is used to manipulate and prepare materials. \textless arg1\textgreater refers to the primary material to be kneaded, such as dough or clay. \textless arg2\textgreater is optional and denotes the surface or item against which the kneading is performed, like a tray or a rolling board.',\\
    `stop': `stop(\textless arg1\textgreater, \textless arg2\textgreater) is used to terminate an ongoing process. \textless arg1\textgreater refers to the type of process being stopped, such as 'liquid'. \textless arg2\textgreater is optional and specifies the equipment involved, like 'gas cooker'. If \textless arg2\textgreater is not provided, the function defaults to stopping all processes related to \textless arg1\textgreater.',\\
    `cook': `cook(\textless arg1\textgreater, \textless arg2\textgreater) is used to simulate the cooking process of a specified ingredient with or without a utensil. \textless arg1\textgreater refers to the item to be cooked, which is a mandatory argument. \textless arg2\textgreater is optional and specifies the tool used in the cooking process, defaulting to none if not provided.',\\
    `rest': `rest(\textless arg1\textgreater, \textless arg2\textgreater) is used to model the passive placement of one object against or on another. \textless arg1\textgreater refers to the primary object that is being supported or placed, while \textless arg2\textgreater is optional and refers to the object or surface against which \textless arg1\textgreater is resting. If \textless arg2\textgreater is omitted, the function defaults to a predetermined resting position or surface.',\\
    `increase\_temperature': `increase\_temperature(\textless arg1\textgreater, \textless arg2\textgreater) is used to raise the temperature of a device using a control mechanism. \textless arg1\textgreater refers to the device whose temperature needs to be increased, such as a cooker or heater. \textless arg2\textgreater is optional and refers to the specific method or interface, like a control knob or button, used to increase the temperature; if not specified, a default method is used.',\\
    `dab': `dab(\textless arg1\textgreater, \textless arg2\textgreater) is used to absorb or blot excess liquid or substances from items. \textless arg1\textgreater refers to the object that requires dabbing, while \textless arg2\textgreater is optional and specifies the material used for dabbing. If \textless arg2\textgreater is omitted, a standard method of dabbing is applied.',\\
    `fix': `fix(\textless arg1\textgreater, \textless arg2\textgreater) is used to attach or affix \textless arg1\textgreater to \textless arg2\textgreater. \textless arg1\textgreater refers to the object or component that needs to be fixed, while \textless arg2\textgreater is optional and identifies the target object or location to which \textless arg1\textgreater will be attached. If \textless arg2\textgreater is omitted, \textless arg1\textgreater is attached to a default object or location.',\\
    `dry': `dry(\textless arg1\textgreater, \textless arg2\textgreater) is used to remove moisture from specified items. \textless arg1\textgreater refers to the item needing drying, like "hands" or "mango." \textless arg2\textgreater is optional and indicates the material used to aid the drying, such as "towel" or "napkin."',\\
    `hang': `hang(\textless arg1\textgreater, \textless arg2\textgreater) is used to place an object onto a specified or default location for storage or accessibility. \textless arg1\textgreater refers to the object to be hung, and \textless arg2\textgreater is optional and denotes the location where the object should be placed. If \textless arg2\textgreater is omitted, a default location is used.',\\
    `tie': `tie(\textless arg1\textgreater, \textless arg2\textgreater) is used to wrap or secure items. \textless arg1\textgreater refers to the material used for tying, such as strings or wraps. \textless arg2\textgreater is optional and indicates additional materials or conditions that might affect the tying process, such as environmental factors or secondary materials.',\\
    `sprinkle': `sprinkle(\textless arg1\textgreater, \textless arg2\textgreater) is used to apply a substance over a surface or object. \textless arg1\textgreater refers to the material to be sprinkled, which is mandatory. \textless arg2\textgreater is optional and defines the surface or object where \textless arg1\textgreater is to be applied. If \textless arg2\textgreater is omitted, the substance is applied to a default location.',\\
    `swing': `swing(\textless arg1\textgreater) is used to alter or move an object in a predefined manner. \textless arg1\textgreater refers to the object being manipulated and the specific actions depend on the nature of this object.',\\
    `fill': `fill(\textless arg1\textgreater, \textless arg2\textgreater) is used to insert a specified substance into a designated container. \textless arg1\textgreater refers to the container that will contain the substance, and \textless arg2\textgreater describes the substance to be filled into the container.',\\
    `wear': `wear(\textless arg1\textgreater, \textless arg2\textgreater, \textless arg3\textgreater) is used to simulate the action of dressing a character or entity with a specific item. \textless arg1\textgreater is mandatory and refers to the item to be worn, described as a string. \textless arg2\textgreater and \textless arg3\textgreater are optional, allowing for customization of style and size, respectively.',\\
    `unsure': `unsure(\textless arg1\textgreater, \textless arg2\textgreater, etc) is used to perform an ambiguous action based on the provided context or data. \textless arg1\textgreater is a mandatory parameter that provides the necessary context or data for the operation of the function. \textless arg2\textgreater and other additional arguments are optional and enhance the function's flexibility and adaptability to varying use cases.',\\
    `sort': `sort(\textless arg1\textgreater, \textless arg2\textgreater) is used to organize or prioritize items based on specific criteria. \textless arg1\textgreater is mandatory and specifies the operation to be performed, while \textless arg2\textgreater is optional and includes the items to be sorted. This function can be used with varying numbers of arguments to adapt to different sorting requirements or settings.',\\
    `stretch': `stretch(\textless arg1\textgreater) is used to modify the physical state of a malleable material by elongating or thinning it. \textless arg1\textgreater refers to the malleable material that is altered by the function.',\\
    `squeeze': `squeeze(\textless arg1\textgreater, \textless arg2\textgreater, etc) is used to compress or reduce the size of various types of input objects. \textless arg1\textgreater refers to the object or substance to be compressed. \textless arg2\textgreater and other optional arguments can be added to modify the function based on the specifics of the compression or the context in which it is applied.',\\
    `flatten': `flatten(\textless arg1\textgreater, \textless arg2\textgreater) is used to press and spread a material into a flatter shape. \textless arg1\textgreater is mandatory and specifies the material to be flattened, while \textless arg2\textgreater is optional and represents a tool used to assist in the flattening process. This function is generally used when a uniform thickness is desired or to prepare the material for further processing.',\\
    `climb': `climb(\textless arg1\textgreater) is used to simulate or command an entity to ascend or mount a specified target. \textless arg1\textgreater refers to the object or location that the entity should climb onto.',\\
    `interact': `interact(\textless arg1\textgreater, \textless arg2\textgreater) is used to perform interactions with various entities or objects. \textless arg1\textgreater refers to the entity or object to interact with, which is mandatory. \textless arg2\textgreater is optional and specifies the method or type of interaction desired; if omitted, it defaults to a standard interaction mode.'\}\\
\textnormal{[end of actionable function and documentation]}\\
Based on the low-level actionable actions provided, you will need to make one or more function calls in order to achieve the mid-level step described in the question.\\
Respond needs to strictly be a list of these actionable functions following this format: ``fuction1(args)'',``fuction2(args)'',``fuction3(args)'', ...\\
Besides these functions, your response should not contain anything else,these functions should not be numbered or explained, simply separated by commas and output directly.\\
For example: ``put(jar, cabinet)'',``grab(jar)'',``mix(jar)'',``put(jar, cabinet)''.\\
You should not include any other text in your response.\\
Question: \{question\}\\
List of actionable functions:}

    \item \textbf{Object grounding in visual grounding:} \textit{\{question\} Please give out the bounding box coordinates of the object.}
    \item \textbf{Frame grounding in visual grounding:} \textit{\{question\} Analyze the provided video and identify the frame number of the last keyframe that is relevant to the specified query. Please provide only the frame number as your response.}
    \item \textbf{Temporal grounding in visual grounding:}
    \textit{\{question\} Please provide the starting and ending times for that step.}
    \item \textbf{Critique in reward modeling:} \textit{Imagine you are the camera wearer (I) who recorded the video. Please directly answer yes or no to determin whether the task is completed or not. Question: \{question\} Short answer:}
    \item \textbf{Feedback in reward modeling:} \textit{Imagine you are the camera wearer (I) who recorded the video. The video contains an uncompleted task. Please identify the essential completion signals in my observations that indicate the task is not completed by me. Please directly generate the rationale as short as possible. Question: \{question\} Short Answer:}
\end{itemize}

\subsection{Evaluation Prompts}

Here we list the prompts for API-based models to assess the performance for some tasks.

\begin{itemize}[leftmargin=*, itemsep=1pt, topsep=0pt]
    \item \textbf{Video question answering:} \textit{[Instruction]\textbackslash nPlease act as an impartial judge and evaluate the quality of the response provided by an AI assistant to the user question displayed below. Your evaluation should consider correctness and helpfulness. You will be given a reference answer and the assistant's answer. Begin your evaluation by comparing the assistant's answer with the reference answer. Identify and correct any mistakes. The assistant has access to an image alongwith questions but you will not be given images. Therefore, please consider only how the answer is close to the reference answer. If the assistant's answer is not exactly same as or similar to the answer, then he must be wrong.  Be as objective as possible. Discourage uninformative answers. Also, equally treat short and long answers and focus on the correctness of answers.  After providing your explanation, you must rate the response with either 0, 0.5 or 1 by strictly following this format:``[[rating]]'', for example: ``Rating: [[0.5]]''.\textbackslash n\textbackslash n[Question]\textbackslash n\{question\}\textbackslash n\textbackslash n[The Start of Reference Answer]\textbackslash n\{ref\_answer\_1\}\textbackslash n[The End of Reference Answer]\textbackslash n\textbackslash n[The Start of Assistant's Answer]\textbackslash n\{answer\}\textbackslash n[The End of Assistant's Answer]''}
    \item \textbf{High-to-mid in hierarchy planning:} \textit{[Instruction]\textbackslash nPlease act as an impartial judge and evaluate the quality of the response provided by an AI assistant to the user question displayed below. Your evaluation should consider correctness and helpfulness. You will be given a reference answer and the assistant's answer. Begin your evaluation by comparing the assistant's answer with the reference answer. Identify and correct any mistakes. The assistant has access to an image alongwith questions but you will not be given images. Therefore, please consider only how the answer is close to the reference answer. The reference answer and the assistant's answer both describe a mid-level step towards completing a high-level goal, you must consider if these two mid-level steps are similar. If the assistant's answer is not exactly same as or similar to the answer, then he must be wrong.  Be as objective as possible. Discourage uninformative answers. Also, equally treat short and long answers and focus on the correctness of answers.  After providing your explanation, you must rate the response with either 0, 0.5 or 1 by strictly following this format: ``[[rating]]'', for example: ``Rating: [[0.5]]\''.\textbackslash n\textbackslash n[Question]\textbackslash n\{question\}\textbackslash n\textbackslash n[The Start of Reference Answer]\textbackslash n\{ref\_answer\_1\}\textbackslash n[The End of Reference Answer]\textbackslash n\textbackslash n[The Start of Assistant's Answer]\textbackslash n\{answer\}\textbackslash n[The End of Assistant's Answer]}
    \item \textbf{Mid-to-low in hierarchy planning:} \textit{[Instruction]\textbackslash nPlease act as an impartial judge and evaluate the quality of the response provided by an AI assistant to the user question displayed below. Your evaluation should consider correctness and helpfulness. You will be given a reference answer and the assistant's answer. Begin your evaluation by comparing the assistant's answer with the reference answer. Identify and correct any mistakes. The assistant has access to an image alongwith questions but you will not be given images. Therefore, please consider only how the answer is close to the reference answer. The reference answer and the assistant's answer both describe a trajectory of low-level automic actions towards completing a mid-level step, you must consider if these two trajectories of low-level atomic actions are similar, especially the key actions to achieve the mid-level step. If the assistant's answer is not exactly same as or similar to the answer, then he must be wrong. Be as objective as possible. After providing your explanation, you must rate the response on a scale of 0 to 10 by strictly following this format: ``[[rating]]'', for example: ``Rating: [[5]]''.\textbackslash n\textbackslash n[Question]\textbackslash n\{question\}\textbackslash n\textbackslash n[The Start of Reference Answer]\textbackslash n\{ref\_answer\_1\}\textbackslash n[The End of Reference Answer]\textbackslash n\textbackslash n[The Start of Assistant's Answer]\textbackslash n\{answer\}\textbackslash n[The End of Assistant's Answer]}
    \item \textbf{Feedback in reward modeling:} \textit{[Instruction]\textbackslash nPlease act as an impartial judge and evaluate the quality of the response provided by an AI assistant to the user question displayed below. Your evaluation should consider correctness and helpfulness. You will be given three reference answers and the assistant's answer. Begin your evaluation by comparing the assistant's answer with the reference answers. Identify and correct any mistakes. The assistant has access to an image alongwith questions but you will not be given images. Therefore, please consider only how the answer is close to the reference answers. If the assistant's answer is not exactly same as or similar to all reference answers, then he must be wrong. If the assistant's answer is exactly same as or similar to any one reference answer, then it is correct. Be as objective as possible. Discourage uninformative answers. Also, equally treat short and long answers and focus on the correctness of answers.  After providing your explanation, you must rate the response with either 0, 0.5 or 1 by strictly following this format: ``[[rating]]'', for example: ``Rating: [[0.5]]''.\textbackslash n\textbackslash n[Question]\textbackslash n\{question\}\textbackslash n\textbackslash n[The Start of Reference Answer]\textbackslash n\{ref\_answer\_1\}\textbackslash n[The End of Reference Answer]\textbackslash n\textbackslash n[The Start of Assistant's Answer]\textbackslash n\{answer\}\textbackslash n[The End of Assistant's Answer]}
\end{itemize}

\clearpage
\section{Case Studies}
\label{app:cases}

% \subsection{Video Question Answering}

\begin{figure}[h]
    \centering
    \includegraphics[width=1\linewidth]{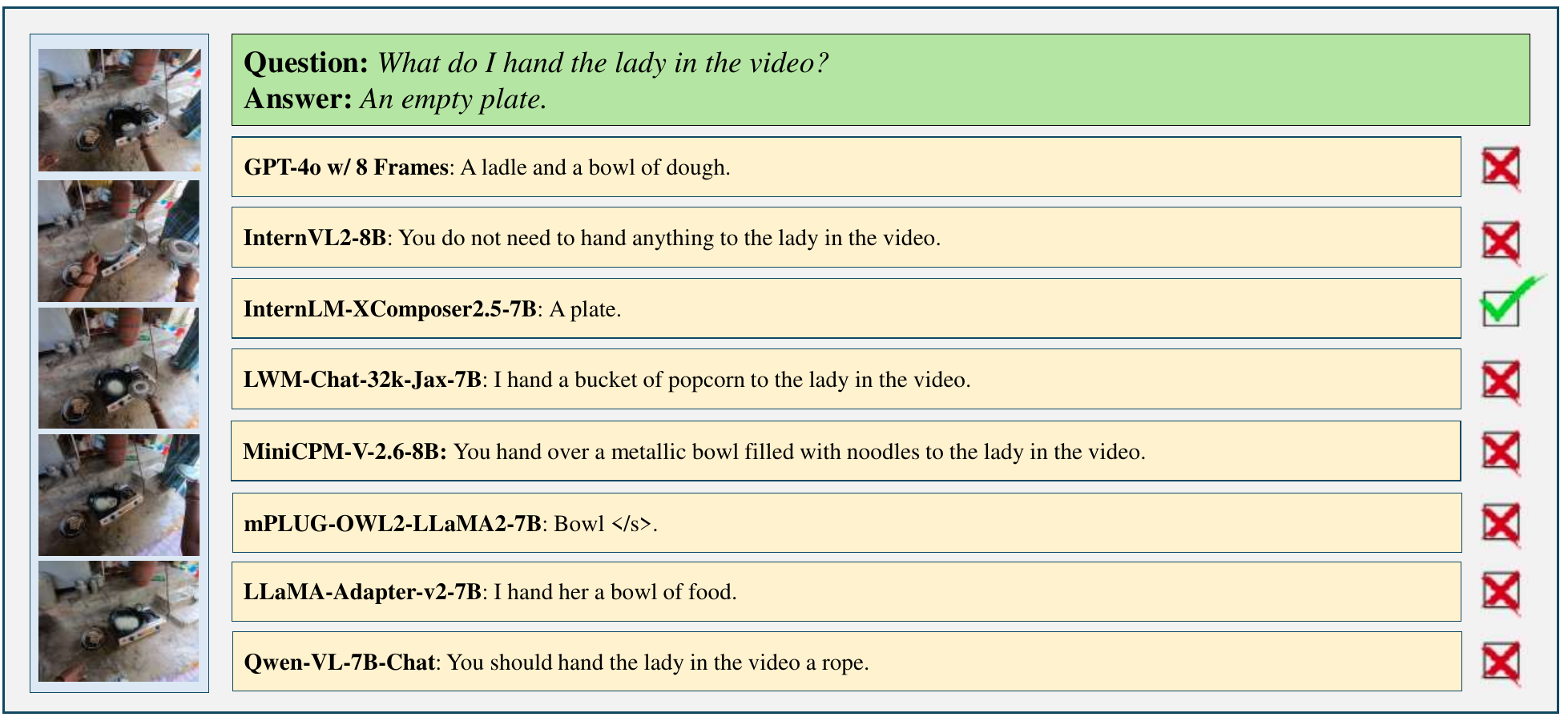}
    \caption{Case of object interaction in video question answering.}
    \label{fig:enter-label}
\end{figure}

\begin{figure}[h!]
    \centering
    \includegraphics[width=1\linewidth]{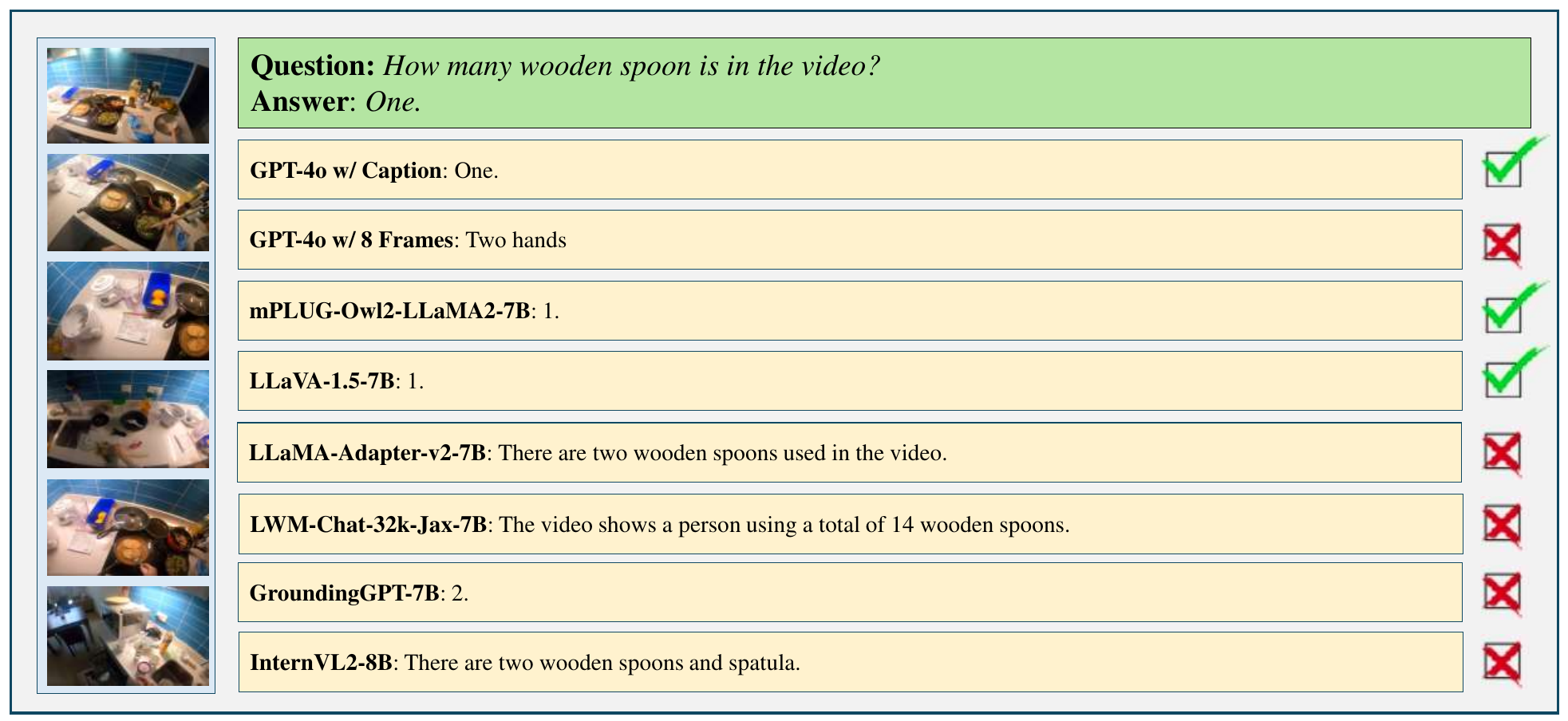}
    \caption{Case of object count of video question answering.}
    \label{fig:enter-label}
\end{figure}

\begin{figure}[h]
    \centering
    \includegraphics[width=1\linewidth]{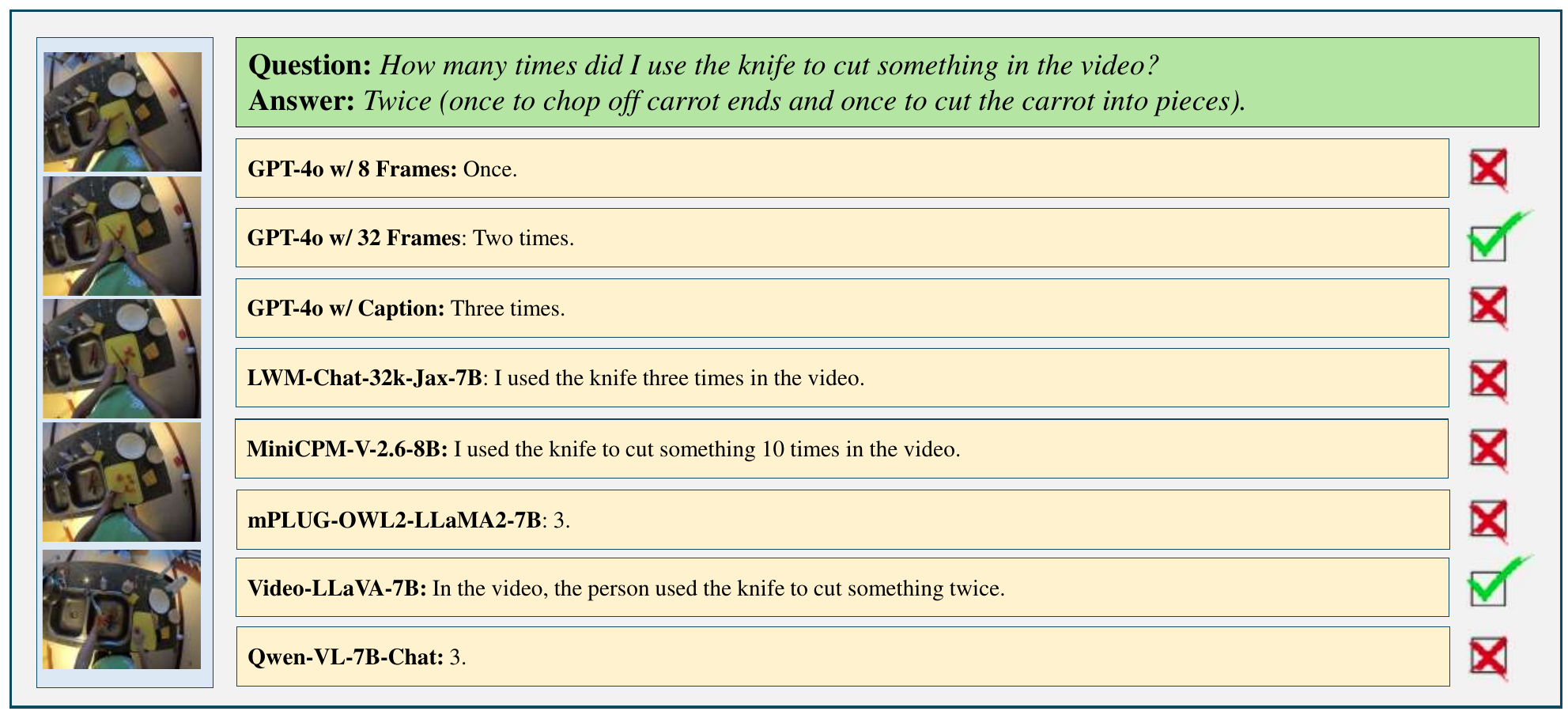}
    \caption{Case of action count of video question answering.}
    \label{fig:enter-label}
\end{figure}

\begin{figure}[h!]
    \centering
    \includegraphics[width=1\linewidth]{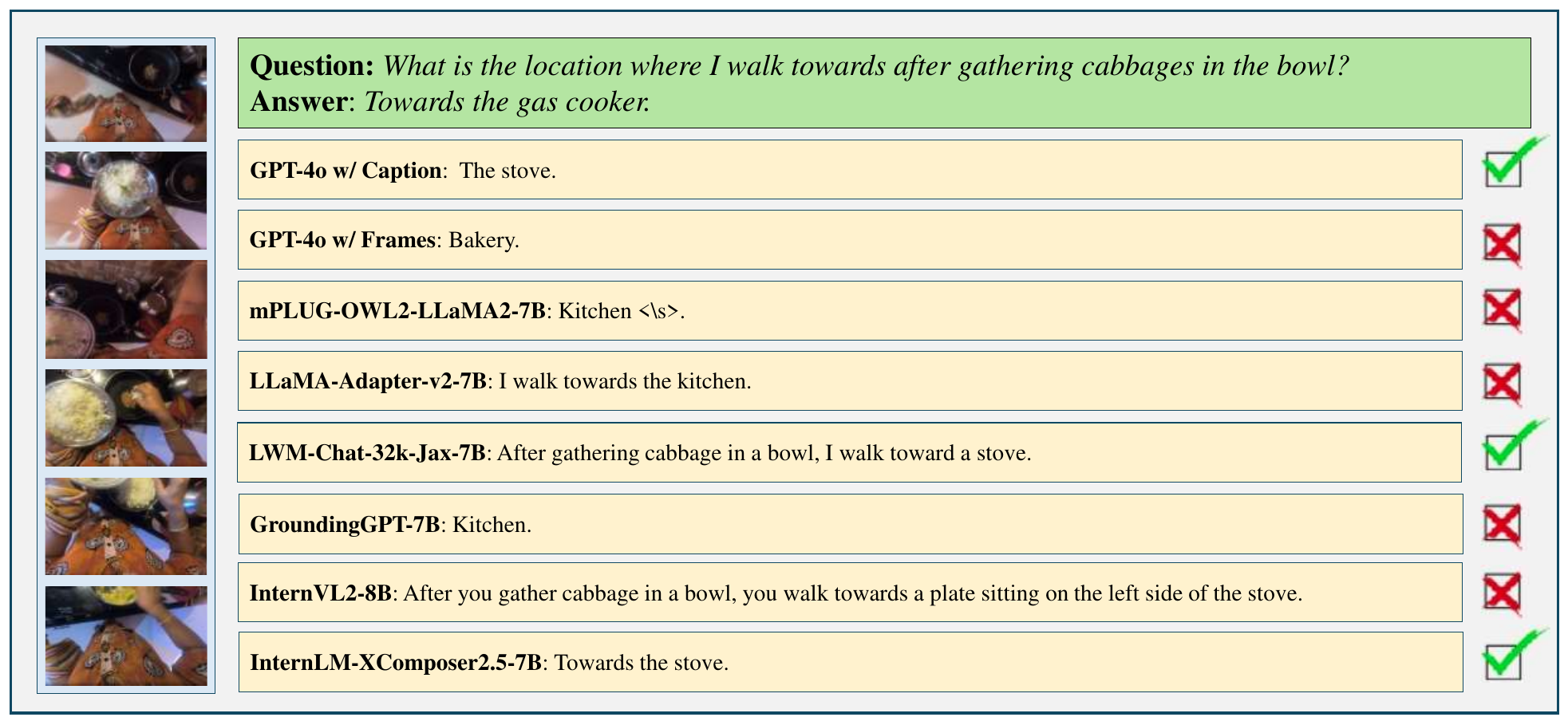}
    \caption{Case of scene transition of video question answering.}
    \label{fig:enter-label}
\end{figure}

% \begin{figure}[h!]
%     \centering
%     \includegraphics[width=1\linewidth]{images/vqa-casestudy-object-ct-ext-cropped.pdf}
%     \caption{Case studies in the visual question and answering dimension: object count (left), object existence (right).}
%     \label{fig:case-hp}
% \end{figure}

% \subsection{Hierarchy Planning}

\begin{figure}[h!]
    \centering
    \includegraphics[width=1\linewidth]{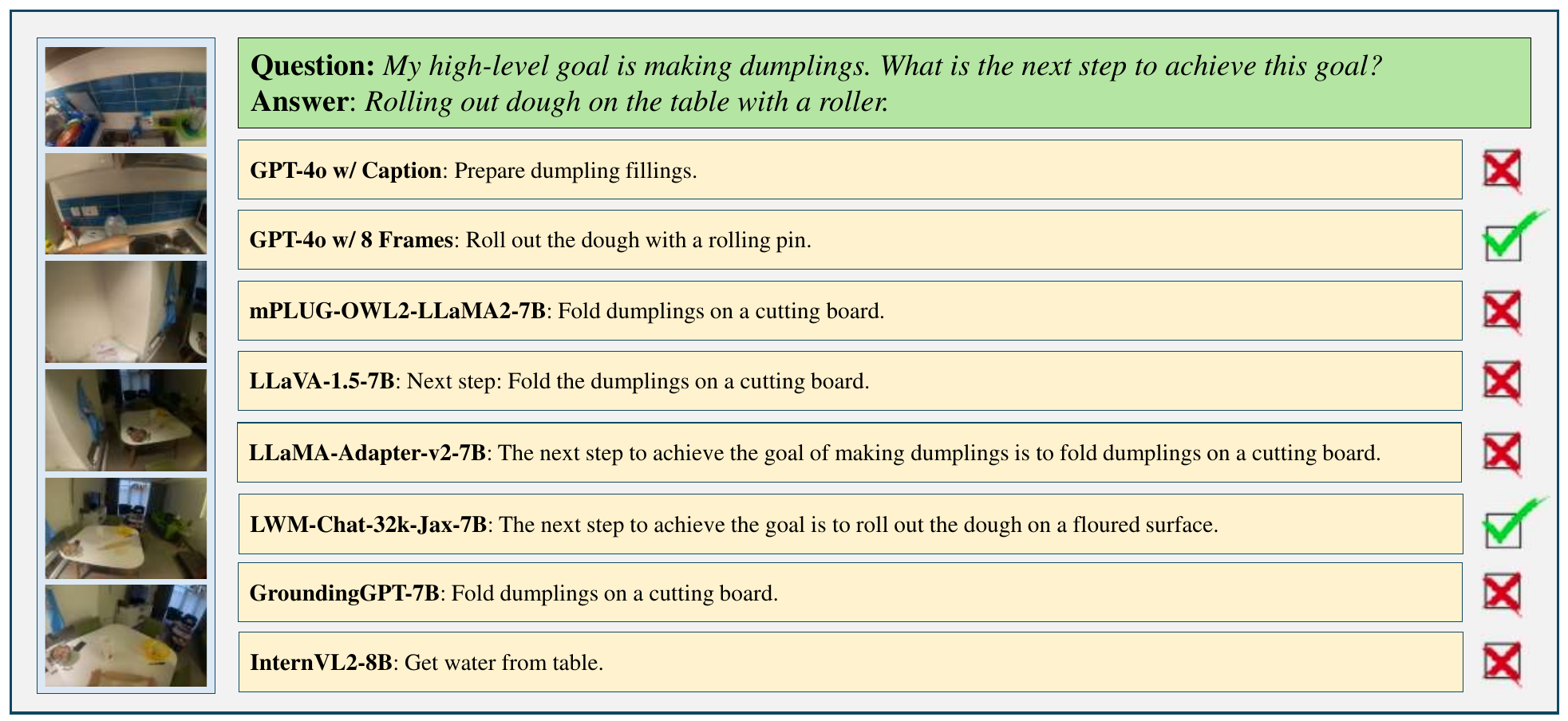}
    \caption{Case of the high-to-mid task in hierarchy planning.}
    \label{fig:enter-label}
\end{figure}

\begin{figure}[h!]
    \centering
    \includegraphics[width=1\linewidth]{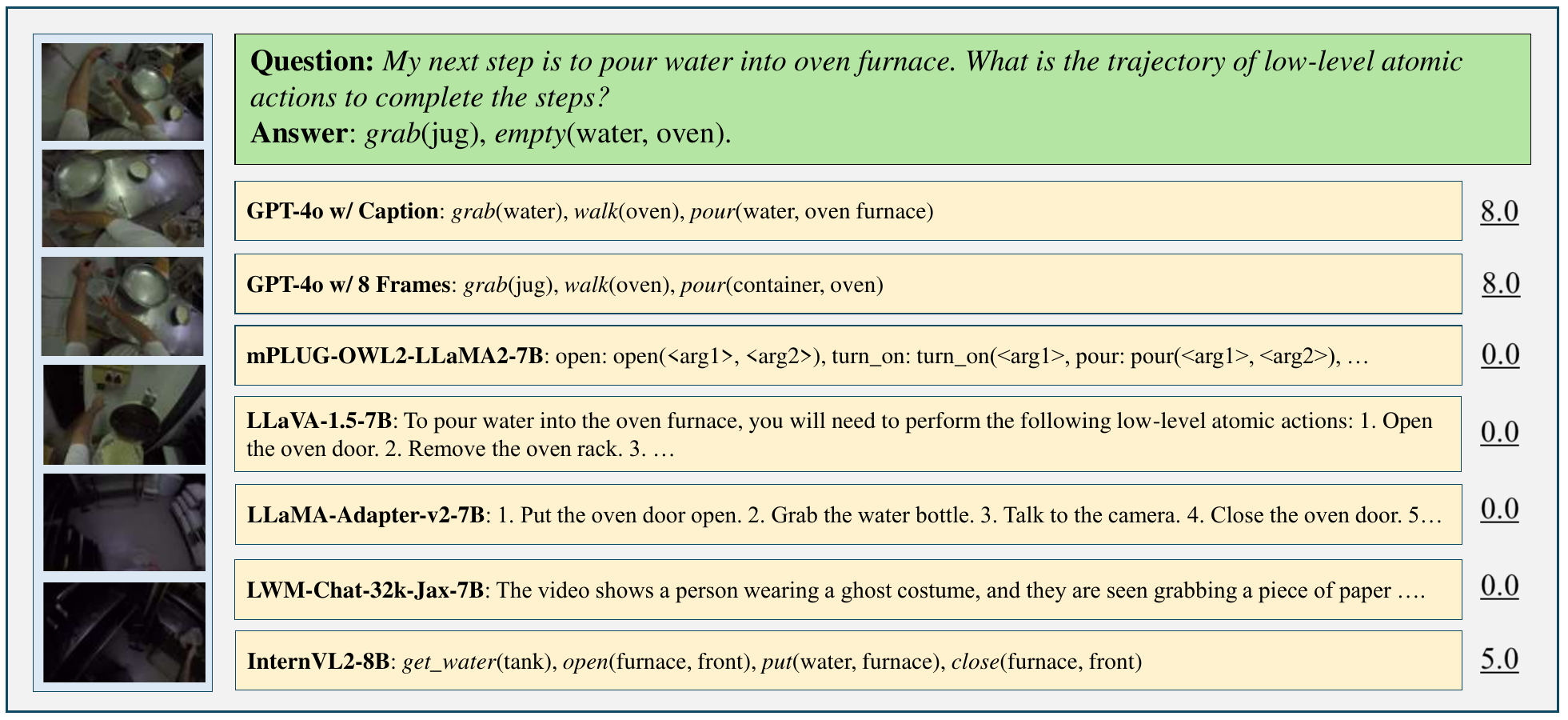}
    \caption{Case of the mid-to-low task in hierarchy planning.}
    \label{fig:enter-label}
\end{figure}

\begin{figure}[h!]
    \centering
    \includegraphics[width=1\linewidth]{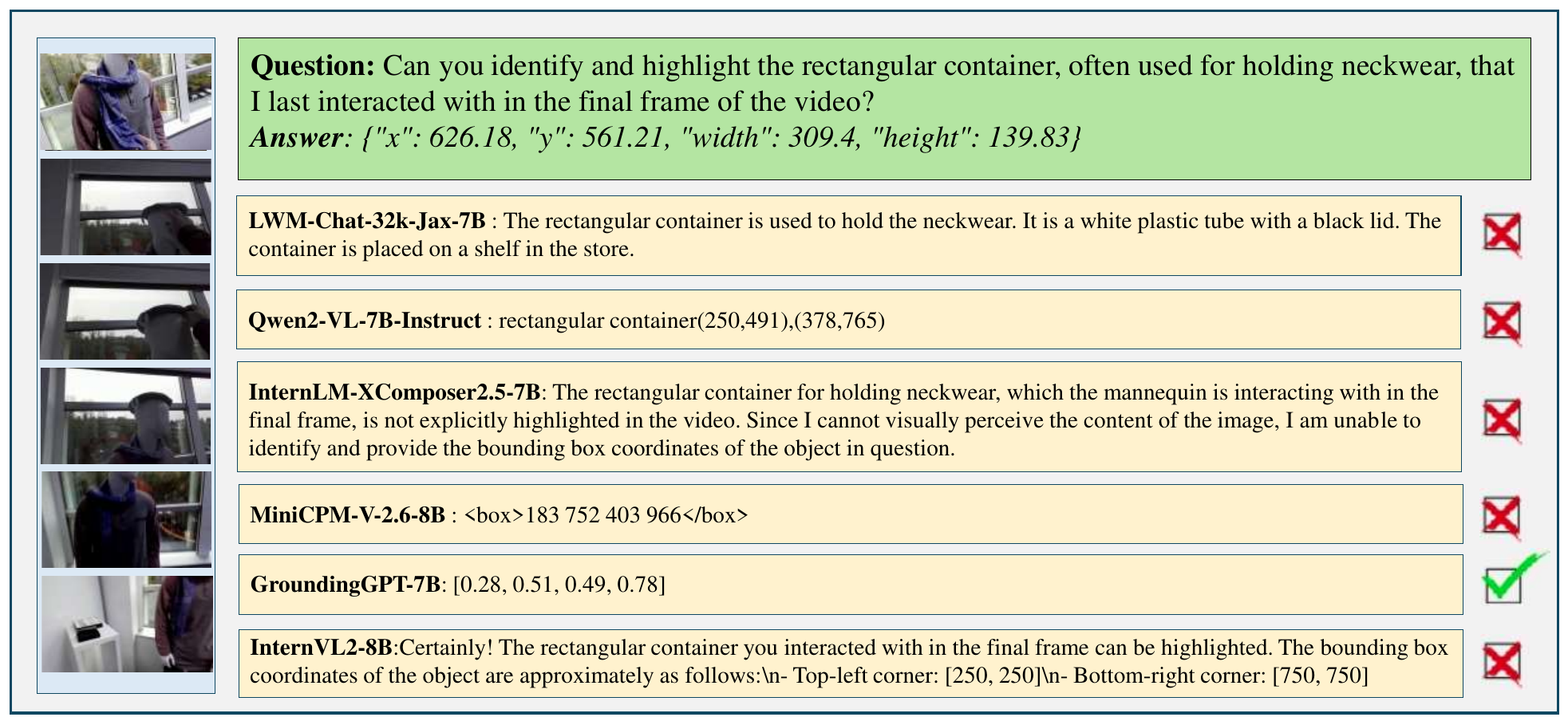}
    \caption{Case of the object grounding in visual grounding. The output of GroundingGPT represents percentage.}
    \label{fig:enter-label}
\end{figure}

\begin{figure}[h!]
    \centering
    \includegraphics[width=1\linewidth]{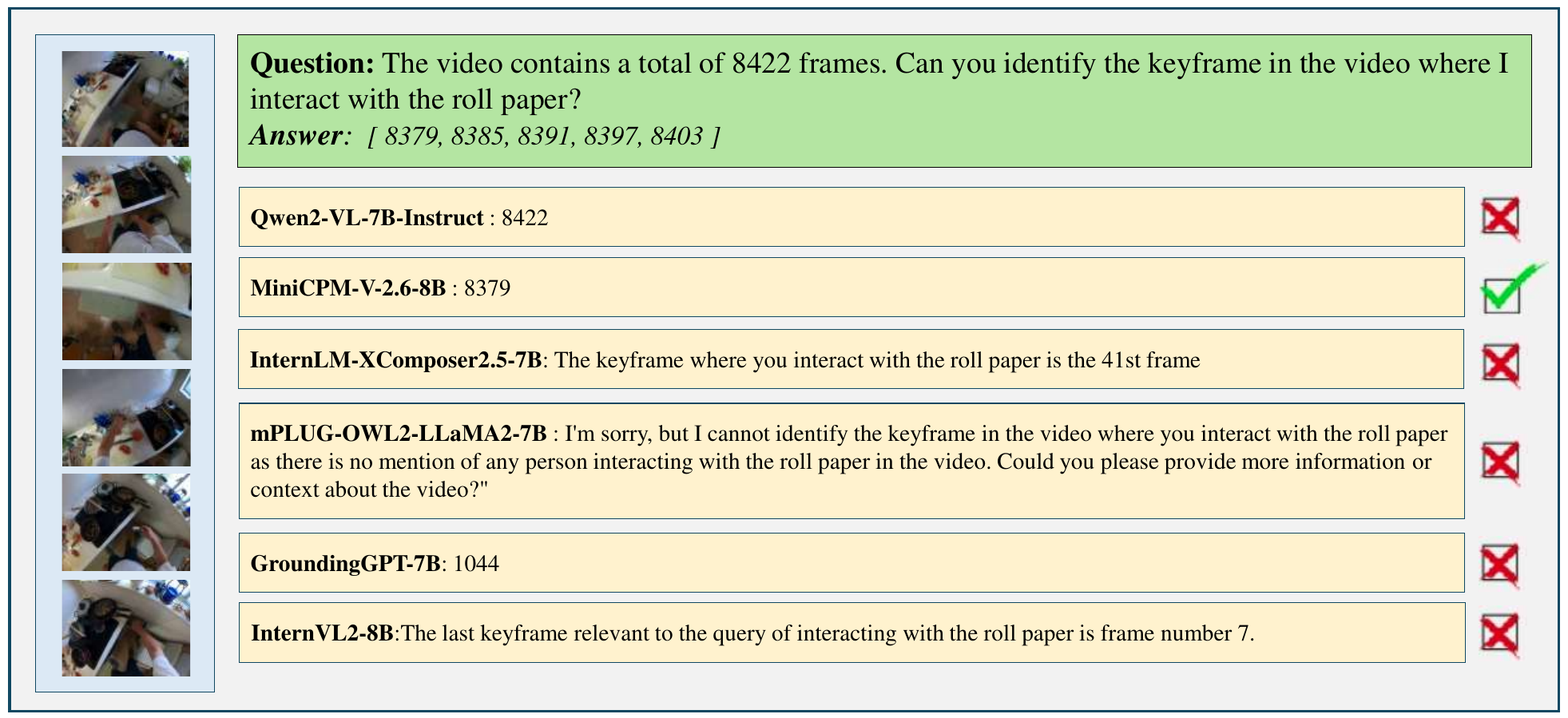}
    \caption{Case of frame grounding in visual grounding.}
    \label{fig:enter-label}
\end{figure}

\begin{figure}[h!]
    \centering
    \includegraphics[width=1\linewidth]{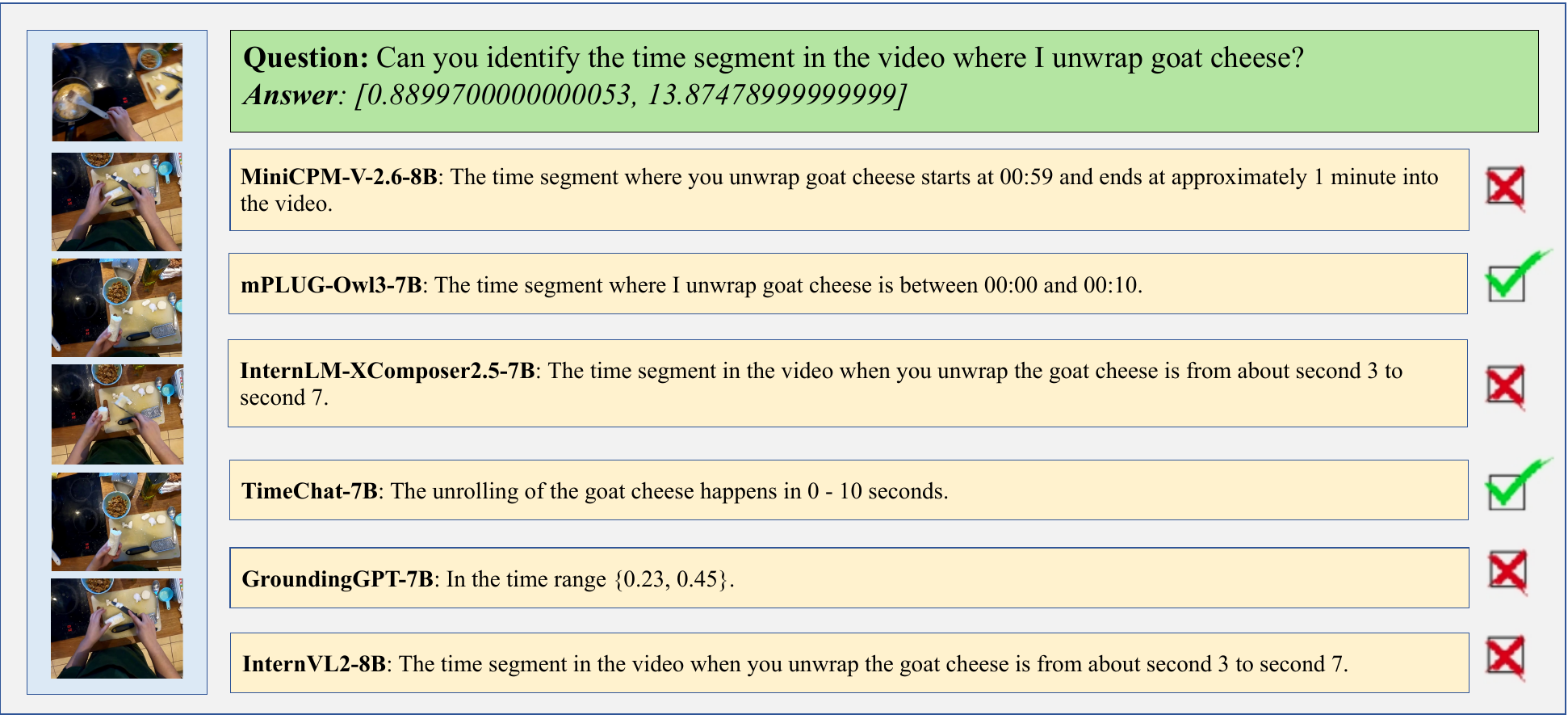}
    \caption{Case of temporal grounding in visual grounding.}
    \label{fig:enter-label}
\end{figure}

% \begin{figure}[h!]
%     \centering
%     \includegraphics[width=1\linewidth]{images/hp-casestudy-figure-cropped.pdf}
%     \caption{Case studies in the hierarchical planning dimension: high to mid (left), mid to low (right).}
%     \label{fig:case-hp}
% \end{figure}

% \subsection{Visual Grounding}

% \subsection{Reward Modeling}

\begin{figure}[h!]
    \centering
    \includegraphics[width=1\linewidth]{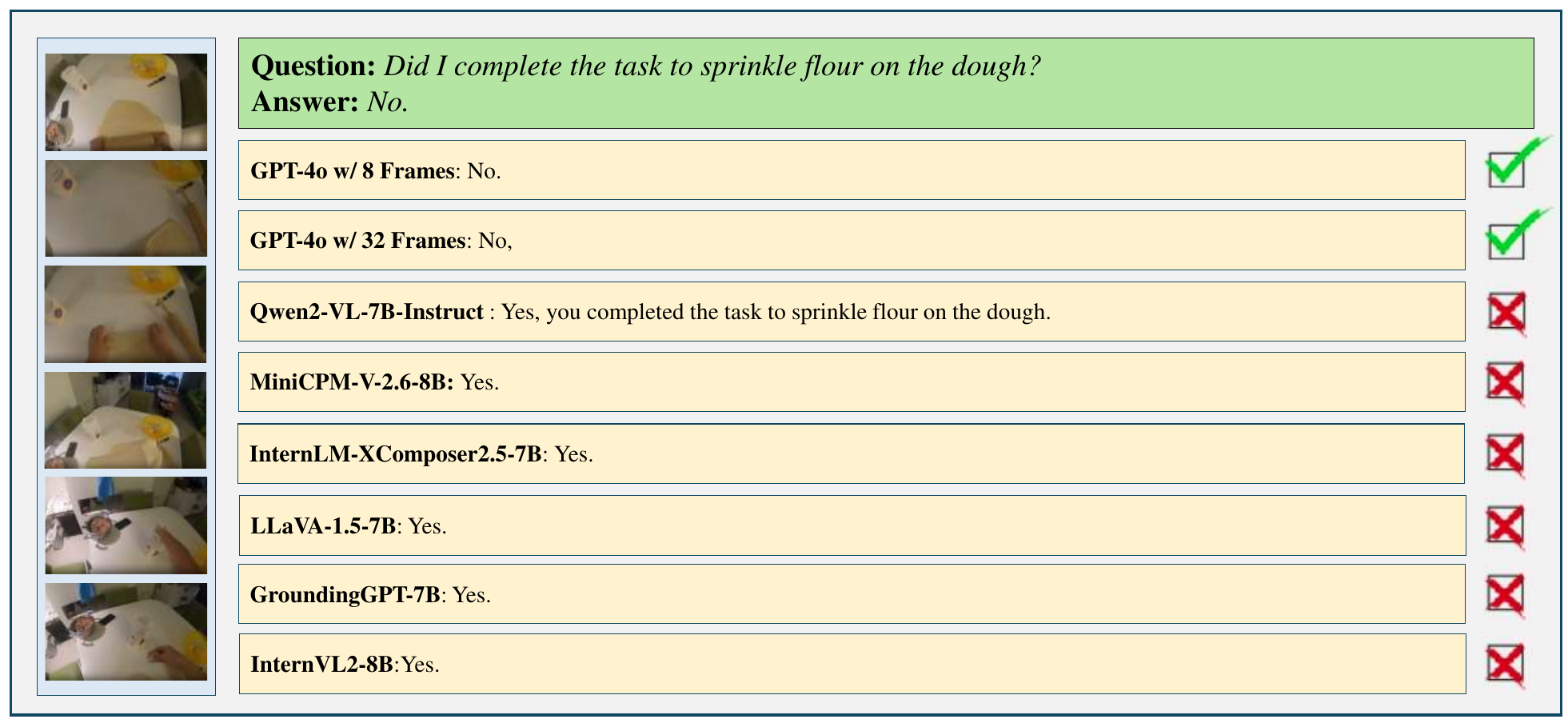}
    \caption{Case of the critique task in reward modeling.}
    \label{fig:enter-label}
\end{figure}

\begin{figure}[h!]
    \centering
    \includegraphics[width=1\linewidth]{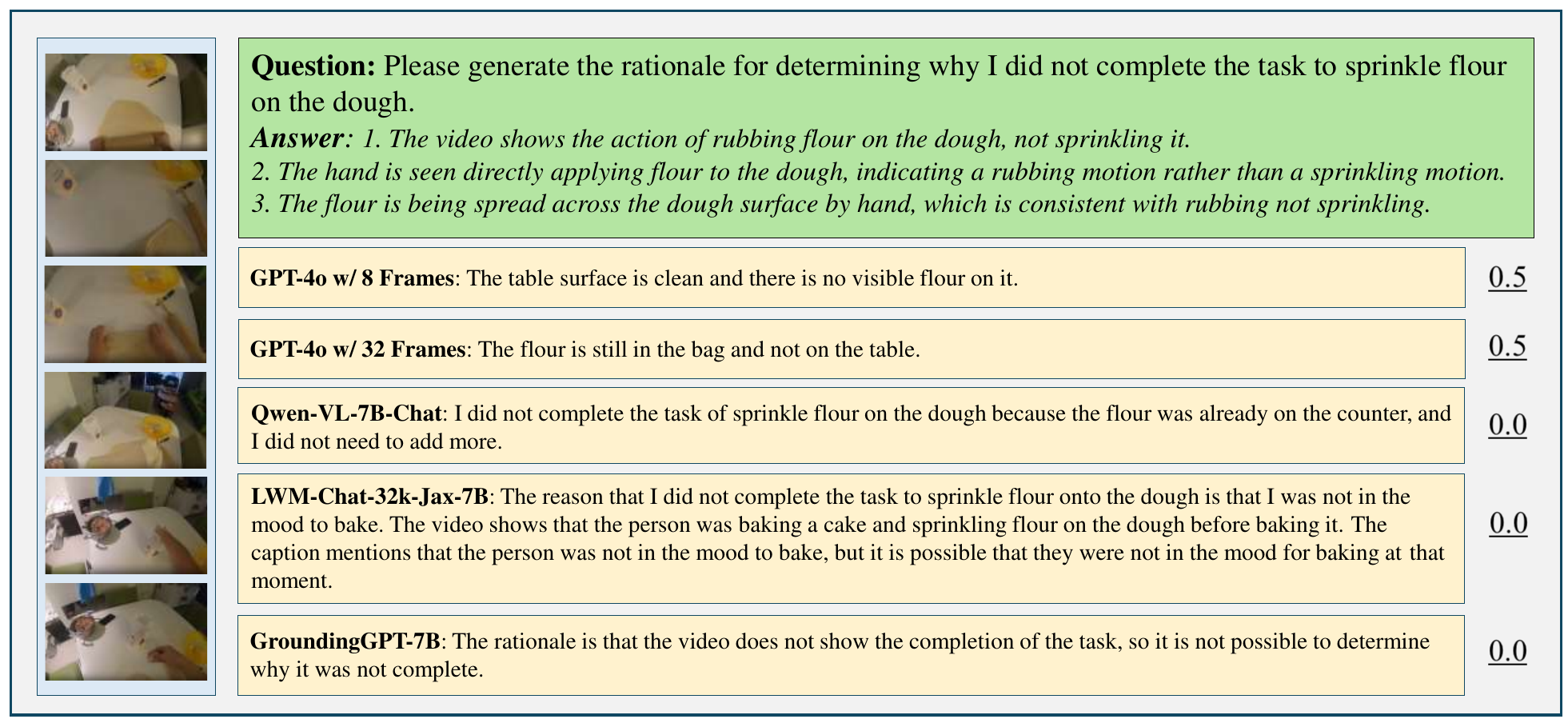}
    \caption{Case of the feedback task in reward modeling.}
    \label{fig:enter-label}
\end{figure}

\end{document}